%% file: 0925_1828SGRL.tex
\documentclass{article} 
\usepackage{iclr2026_conference,times}

\input{math_commands.tex}

\usepackage{hyperref}
\usepackage{url}
\usepackage{graphicx}
\usepackage{threeparttable}
\usepackage{booktabs} 
\usepackage{makecell}
\usepackage{multirow}
\usepackage{tcolorbox}
\usepackage{listings}
\usepackage{xcolor}
\usepackage{subcaption} 
\usepackage{caption} 
\usepackage{wrapfig}
\tcbuselibrary{breakable, skins}

\title{Goal-Guided Efficient Exploration via Large Language Model in Reinforcement Learning}

\author{Yajie Qi, Wei Wei, Lin Li, Lijun Zhang, Zhidong Gao, Da Wang, Huizhong Song\\
School of Computer and Information Technology,
Shanxi University,
Taiyuan, Shanxi, China
}

\iclrfinalcopy 
\begin{document}

\maketitle

\begin{abstract}
Real-world decision-making tasks typically occur in complex and open environments, posing significant challenges to reinforcement learning (RL) agents' exploration efficiency and long-horizon planning capabilities. 
A promising approach is LLM-enhanced RL, which leverages the rich prior knowledge and strong planning capabilities of LLMs to guide RL agents in efficient exploration. 
However, existing methods mostly rely on frequent and costly LLM invocations and suffer from limited performance due to the semantic mismatch. 
In this paper, we introduce a Structured Goal-guided Reinforcement Learning (SGRL) method that integrates a structured goal planner and a goal-conditioned action pruner to guide RL agents toward efficient exploration. 
Specifically, the structured goal planner utilizes LLMs to generate a reusable, structured function for goal generation, in which goals are prioritized.
Furthermore, by utilizing LLMs to determine goals' priority weights,  it dynamically generates forward-looking goals to guide the agent's policy toward more promising decision-making trajectories. 
The goal-conditioned action pruner employs an action masking mechanism that filters out actions misaligned with the current goal, thereby constraining the RL agent to select goal-consistent policies. 
We evaluate the proposed method on Crafter and Craftax-Classic, and experimental results demonstrate that SGRL achieves superior performance compared to existing state-of-the-art methods.
\end{abstract}

\section{Introduction}  \label{Introduction}
Reinforcement learning (RL) has achieved impressive success in addressing challenging decision-making tasks across a wide range of domains, such as Atari games ~\citep{hessel2018rainbow, vinyals2019grandmaster}, robotics ~\citep{brunke2022safe, haarnoja2024learning}, and natural language processing ~\citep{padmakumar2021dialog, ouyang2022training}. 
However, these remarkable successes have mostly occurred in environments characterized by closed, predefined tasks with clear goals and immediate feedback, which fail to capture the complexity and dynamics of the real world~\citep{team2021open, cai2023open}.
In recent years, increasing attention has been devoted to decision-making in open-world environments such as Minecraft~\citep{fan2022minedojo, Lin2022JueWuMCPM} and Crafter~\citep{hafner2022benchmarking, moon2023discovering}, which pose significant challenges in generalization, deep exploration, long-term decision-making, and reasoning.
Consequently, solving decision-making problems in open-world environments is widely recognized as a pressing and significant challenge.

Recently, LLM-enhanced RL has been regarded as a promising direction for addressing decision-making challenges in open-world environments~\citep{liu2023llm, Zhou2024LLM4RL, he2024words, schoepp2025evolving} due to the remarkable capabilities of LLMs in various decision-making and reasoning tasks.
A variety of methods, such as using LLMs as decision-makers~\citep{shinn2023reflexion, carta2023grounding, gaven2024sac} or as skill planners~\citep{ichter2022do, zhang2023bootstrap, lin2023text2motion, yang2025automated}, have emerged and significantly improved sample efficiency and generalization, thereby enhancing performance.
However, these methods still struggle to fully unlock and leverage the planning capabilities of LLMs, as they either force the models to perform fine-grained planning (an area where they are not particularly adept) or fail to effectively coordinate the relationship between the LLM and RL agent components. 
In order to harness the capabilities of LLMs, recent studies have explored a more direct and effective approach that uses an LLM to generate high-level goals for guiding exploration in the RL agent~\citep{du2023guiding,zheng2024online, shukla2024lgts, zhang2024adarefiner}.
To mention a few, ELLM~\citep{du2023guiding} leverages a pre-trained LLM to generate potential exploration goals, guiding the RL agent to explore towards potentially useful targets.
AdaRefiner~\citep{zhang2024adarefiner} iteratively refines task understanding through feedback from the RL agent, thereby improving the quality of LLM-generated goals and fostering more effective collaboration between LLM and RL agent.
However, due to the reliance on frequent and intensive LLM invocations, these methods suffer from low practical utility and poor computational efficiency.

\begin{figure*}[t]
	\centering
	\includegraphics[width=0.98\textwidth]{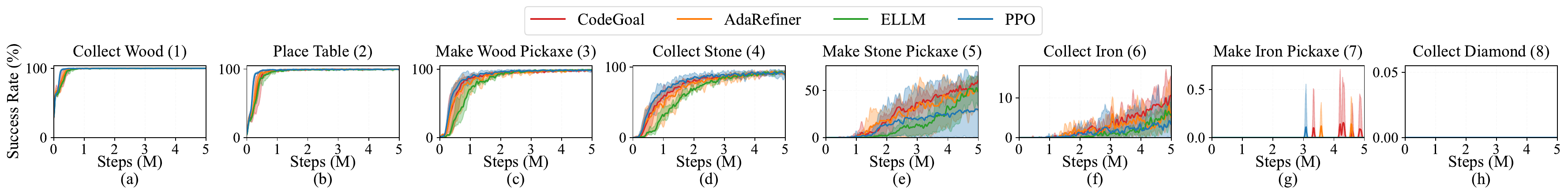} 
	\caption{Success rate curves for 8 main achievements on Craftax-Classic, ordered from left to right by achievement depth from 1 to 8.
	A complete and more intuitive version is shown in Figure \ref{AppendixFig: fig_22ach_tax_mot_5M} of the Appendix~\ref{Appendix C: Additional Preliminary Results}.}
\label{Fig: fig_tax_mot_1x8_5M}
\end{figure*}

Inspired by \citet{ma2024eureka, xie2024text2reward}, in which LLMs generate reward-shaping code to provide immediate feedback to RL agents and achieve strong performance on several benchmarks, we hypothesize that structured, goal-specifying code can also serve as a stable and executable interface between LLMs and RL agents, thereby enabling effective long-horizon exploration. 
Thus, we conducted preliminary experiments based on the simple idea of leveraging LLM-generated code to guide the exploration of RL agents, which we term CodeGoal.
Figure~\ref{Fig: fig_tax_mot_1x8_5M} presents the success rate of CodeGoal in comparison with PPO, as well as {ELLM~\citep{du2023guiding}}\footnote{using DeepSeek-V3 model as goal generator, queried every 200 steps under training time and cost constraints.} and {AdaRefiner~\citep{zhang2024adarefiner}}\footnote{using DeepSeek-V3 model as both adapter and decision LLM, queried every 200 steps under training time and cost constraints.} on the Craftax-Classic benchmark.
From Figure~\ref{Fig: fig_tax_mot_1x8_5M}, we can observe that compared with ELLM~\citep{du2023guiding}, AdaRefiner~\citep{zhang2024adarefiner}, and PPO, CodeGoal achieves competitive success rates when unlocking key achievements, demonstrating that using code-generated goals to guide RL agents is a viable approach.

It is worth noting that the experimental results in Figure~\ref{Fig: fig_tax_mot_1x8_5M} also reveal two key limitations:
(1) the agent fails to quickly unlock deeper achievements such as \textit{Collect Diamond};
(2) integrating textual goals into the RL agent's decision-making process yields limited immediate benefits, evidenced by the apparent ineffectiveness of goal guidance within the first 2M steps. 
This motivates us to propose a novel LLM-based goal-conditioned RL approach, which first leverages LLMs to generate a parameterized, well-structured, and reusable goal-generation function, and then utilizes the LLMs to optimize both the parameters of this function and the goal-conditioned policy constraints, thereby encouraging RL agents to explore effectively in open-world environments.

The main contributions are summarized as follows:
\begin{itemize}
	\item We propose Structured Goal-guided Reinforcement Learning (SGRL), an LLM-enhanced RL method that constructs a structured goal-generation function and dynamically adjusts both goal priority weights and goal-conditioned action constraints, thereby significantly improving the RL agent's exploration efficiency and overall performance, while maintaining low LLM invocation frequency and minimal input token consumption per call.

	\item We develop a structured goal planner that leverages the LLM to construct a reusable, structured goal-generating function that selects forward-looking goals and dynamically adjusts their priority weights during training.
	    Furthermore, a goal-conditioned action pruner is designed to filter out actions misaligned with the goal, thereby constraining agents to select goal-consistent policies. 

	\item Extensive experimental results in the challenging open-world environments Crafter and Craftax-Classic demonstrate that SGRL consistently outperforms or matches existing LLM-enhanced RL methods across multiple metrics, including success rate, total score, cumulative reward, and achievement depth.
	
\end{itemize}

\begin{figure*}[t]
	\centering
	\includegraphics[width=1\textwidth]{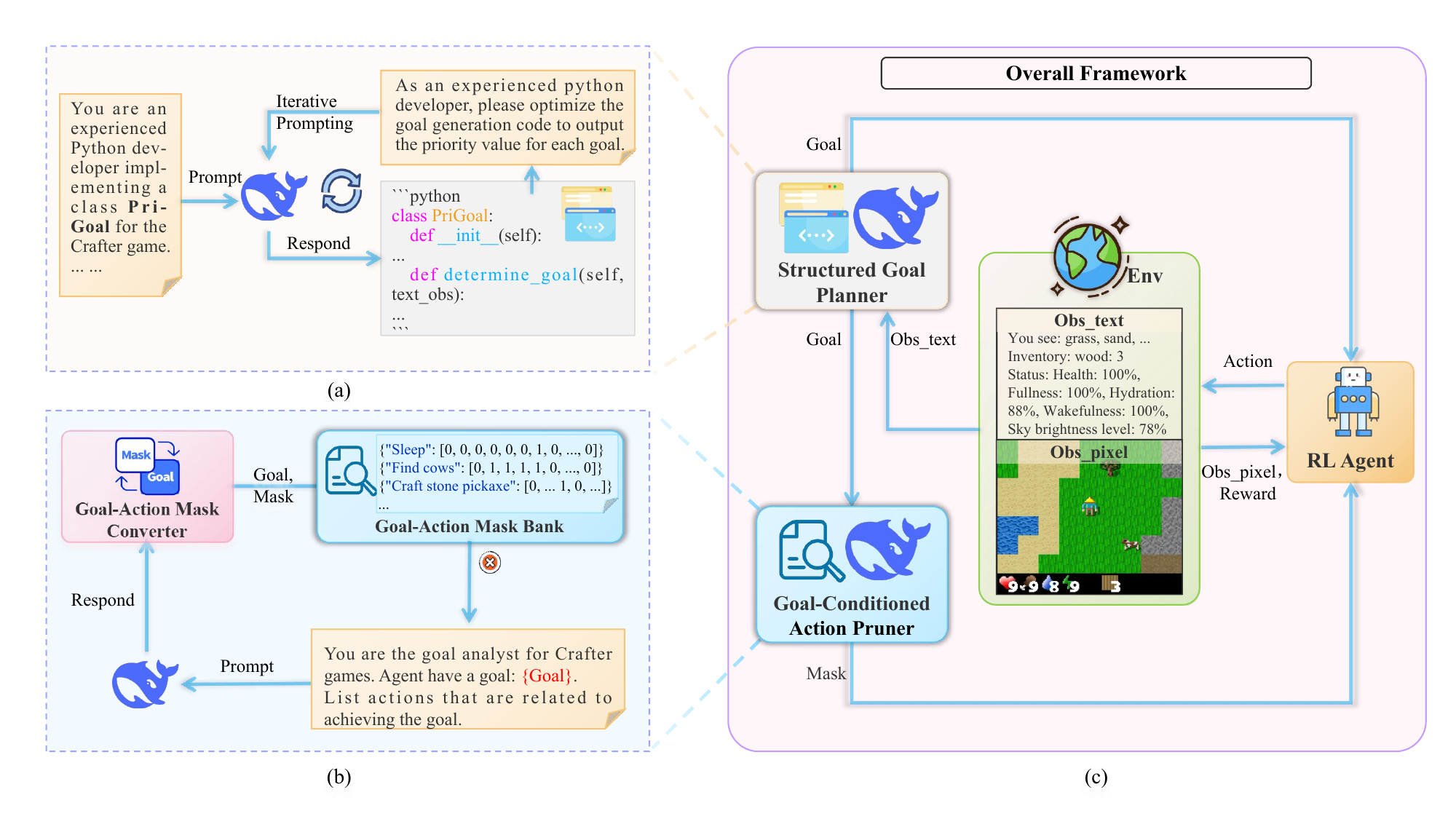} %
	\caption{(a) Structured Goal Planner: generates goals with priority weights based on environment states and distilled task knowledge;
	(b) Goal-Conditioned Action Pruner: filters invalid or irrelevant actions based on current goals;
	(c) The overall framework of SGRL.}
	\label{Fig: framework}
\end{figure*}

\section{Preliminary}
\subsection{Goal-Conditioned Reinforcement Learning}
A goal-conditioned reinforcement learning (GCRL)~\citep{Liu2022Goal} problem can be formulated as a goal-augmented MDP, denoted as a tuple
$<\mathcal{S}, \mathcal{A}, \mathcal{G}, p_g, \phi,\mathcal{P}, r,\rho_0, \gamma>$,
where $\mathcal{S}$, $\mathcal{A}$ and $\mathcal{G}$ denote the state space, action space and goal space, respectively;
$p_g$ and $\rho_0$ denote the desired goal distribution and initial state distribution, respectively;
$\phi: \mathcal{S} \to \mathcal{G}$ is a mapping function that maps the state to a specific goal;
$\mathcal{P}: \mathcal{S} \times \mathcal{A} \to \mathcal{S}$ represents the state transition function;
$r: \mathcal{S} \times \mathcal{A} \times \mathcal{G} \to \mathbb{R}$ is the reward function;
and $\gamma$ is the discount factor.

At each timestep $t$, given the state $s_t$ and the desired goal $g$, the agent takes an action $a_{t}$ according to its policy $\pi(a_t \mid s_t, g)$ to interact with the environment and then receives the next state $s_{t+1}$ and reward $r_{t+1}$.
The agent's goal is to maximize the expected sum of discounted rewards over the state-goal distribution:
\begin{equation}
		\mathcal{J}(\pi)=\mathbb{E}_{\substack{g \sim p_g, s_0 \sim \rho_0, \\ a_t \sim \pi\left(\cdot \mid s_t, g\right)}}\left[\sum_{t} \gamma^{t} r\left(s_{t}, g, a_{t}\right)\right].
\end{equation}

\subsection{LLM as Goal Planner}
Formally, given a natural language task description $\omega \in \Omega$, a sequence of goals is generated by an LLM-based goal planner.
At each planning step $h$, the LLM takes the state-goal history $\tau_h = \{s_1, g_1, \dots, s_{h-1}, g_{h-1}, s_h\}$ as input and generates the next goal:
$$g_h \sim \pi_{\mathrm{LLM}}(\cdot \mid \tau_h, \omega),$$
where the goal $g_h$ is expressed in natural language and belongs to the language space $\mathfrak{L}$, which contains possible linguistic expressions such as sentences or phrases describing actionable intentions. 
For practical purposes, a task-specific subset $\mathcal{G} \subseteq \mathfrak{L}$ is often considered to restrict the goal space to valid and executable instructions.
Subsequently, the agent choose a policy $\pi$ conditioned on both the current environment state $s_h \in \mathcal{S}$ and the goal $g_h$:
$$a_h \sim \pi(\cdot \mid s_h, g_h),$$
where $a_h \in \mathcal{A}$ denotes the agent's action at time step $h$.

\section{Method} \label{Section: Method}
This section develops a novel LLM-enhanced RL method called Structured Goal-guided Reinforcement Learning (SGRL).
Figure~\ref{Fig: framework} (c) shows an overview of the SGRL framework, which consists of three key components: 
(1) a structured goal planner, which receives textual observations from the environment and provides forward-looking goals; 
(2) a goal-conditioned action pruner, which takes the goal from the structured goal planner as input and generates an action mask;
and (3) an RL agent, which executes actions according to a policy conditioned on the current environmental state, goal and most recent reward.
The details are introduced in the next subsections.

\subsection{Structured Goal Planner}
The core task of the structured goal planner is to generate the goal function and optimize the goal priority weights by LLMs through task-specific prompting, as illustrated in Figure~\ref{Fig: framework} (a).
Specifically, given a basic task introduction, its rules, and a few code examples, the LLM generates an executable, structured function for goal generation according to task-specific and environmental state-related prompting.
Then, in just a few iterations and optimizations, we can obtain a reusable, structured goal generation function with priority weights, which can provide forward-looking goals to guide RL agent exploration efficiency. 
Furthermore, in actual execution, the structured goal planner adjusts the weights of goals based on the agent's unlocked achievements at different training stages.

Formally, at each timestep $t$, the structured goal planner constructs a function $\{(g^i_t, w^i_t)\}_{i=1}^k \sim \phi(s_t)$, which maps the current state $s_t \in \mathcal{S}$ to a set of $k$ candidate goals $g_t$ with priority weights $w_t$. 
Here, $g^i_t \in \mathcal{G}$ denotes the $i$-th goal from the goal space $\mathcal{G}$, and $w^i_t \in [0, 1]$ is the normalized priority value for that goal, satisfying $\sum_{i=1}^k w_i = 1$. 
These priority values serve to rank candidate goals to guide the agent's exploration toward the most relevant and achievable objectives. 
The goal planning function $\phi(s)$ is constructed directly by LLM with task-specific and current state-related prompts, unlike the method in which the LLM as a goal planner that generates goals $g_t \sim \pi_{\mathrm{LLM}}(\cdot \mid \tau_t, \omega)$ without explicitly defining a reusable goal-generating function.
It is worth noting that the structured goal generation method can provide semantically forward-looking goals to improve the RL agent's exploration efficiency, while maintaining a low LLM invocation frequency and minimal input token consumption per call.

\subsection{Goal-Conditioned Action Pruner}
The core task of the goal-conditioned action pruner is to constrain the agent to select goal-consistent policies, as shown in Figure~\ref{Fig: framework}~(b).
Specifically, we first leverage the LLM with rich prior knowledge to filter from the candidate action set those actions aligned with the current goal. 
This filtering is implemented via a goal-action masking mechanism.
In practice, we construct a goal-action mask bank to store goal-mask pairs, eliminating redundant LLM queries when the same goal reappears.

Formally, at each timestep $t$, the action pruner produces a binary mask $m(g^i_t) \in {0,1}^{|\mathcal{A}|}$ base on the candidate goals $\{g^1_t, \dots, g^k_t\}$, which are obtained from the LLM or the goal-action mask bank.
Each element of $m(g^i_t)$ indicates whether the action is relevant to the goal $g^i_t$.
Then, the action pruner extracts actions relevant to any given goal $g^i_t$ in an element-wise fashion as follows:
\begin{equation}
	M = \max\nolimits_{i=1, \dots, k} \, m(g^i_t).
\end{equation}

However, it is worth noting that strictly adhering to goal guidance at all times is not always optimal, especially when using a general-purpose LLM without fine-tuning or domain-specific expertise. 
Therefore, we introduce a masking coefficient that grants the RL agent some autonomy in decision-making.
Specifically, we design a three-phase cosine annealing schedule for the exploration coefficient $\xi \in [0,1]$, which gradually adjusts the strength of action masking throughout training. Specifically:
\begin{equation}\renewcommand{\arraystretch}{1.5}
    \xi(t)=\left\{\begin{array}{ll}
        \frac{1}{2}\left(1+\cos \left(\frac{t}{0.4 T} \cdot \pi \right)\right),       & 0 \leq t<0.4 T     \\
        \frac{1}{2}\left(1-\cos \left(\frac{t-0.4 T}{0.4 T} \cdot \pi \right)\right), & 0.4 T \leq t<0.8 T \\
        1.0,                                                                        & t \geq 0.8 T
    \end{array}\right.
    .
    \label{eq-epsilon}
\end{equation}
Then, the coefficient $\xi$ is used to stochastically relax the mask via a Bernoulli sampling mechanism:
\begin{equation} \label{Equation: Bernoulli sampling}
    \tilde{M}^j = \max\left(M^j, \text{Bernoulli}(\xi \cdot (1 - M^j))\right),
\end{equation}
where $M^j$ is the the $j-th$ element of $M$, $j = 1,\dots,|\mathcal{A}|$.
Equation (\ref{Equation: Bernoulli sampling}) allows masked-out actions to be sampled with a small probability, thereby mitigating policy rigidity caused by inaccurate masks or shifts in environmental dynamics.
Notably, we provide the performance comparison of algorithms with different masking strategies in Appendix~\ref{Appendix F: Additional Pruner Epsilon Annealing Strategies and Implementation}.

\subsection{RL Agent}
In this subsection, we introduce how the RL agent makes decisions based on its current state, as well as goals and action mask constraints.

First, the goal set with priority weights $\{(g^i_t, w^i_t)\}_{i=1}^k$ is encoded into a goal embedding vector $g_t^{\text{emb}}$ via the encoder network. 
The embedding vector $g_t^{\text{emb}}$ is concatenated with the state $s_t$ to form an augmented state $[s_t; g_t^{\text{emb}}]$, which serves as the input to the policy network $\pi_\theta(\cdot \mid s_t, g_t^{\text{emb}})$.
Then, based on the action mask $\tilde{M}$, the raw logits output can be obtained to enforce action feasibility: 
\begin{equation}
    \texttt{logits} = \texttt{actor\_logits} \odot \tilde{M} + (1 - \tilde{M}) \cdot (-C),
\end{equation}
where $\odot$ denotes element-wise multiplication, and $C \gg 0$ is a sufficiently large constant (e.g., $10^6$) that suppresses the logits of invalid actions to near negative infinity.

In practice, the policy $\pi_\theta(a_t \mid s_t, g_t^{\text{emb}})$ of the RL agent is updated using the PPO algorithm~\citep{schulman2017proximal} with state augmentation by optimizing the following objective:
\begin{equation}
	\begin{aligned}
		\mathcal{L}_\pi = \mathbb{E}_{\substack{
			s_t, g_t^{\text{emb}} \sim \mathcal{D}, \\
			a_t \sim \pi_{\text{old}}(\cdot \mid s_t, g_t^{\text{emb}}), \\
			s_{t+1} \sim \mathcal{P}(\cdot \mid s_t, a_t)
		}} \Bigg[\min\Bigg( 
			\frac{\pi_\theta(a_t \mid s_t, g_t^{\text{emb}})}{\pi_{\text{old}}(a_t \mid s_t, g_t^{\text{emb}})} \hat{A}_t, 
		\text{clip}\left( \frac{\pi_\theta(a_t \mid s_t, g_t^{\text{emb}})}{\pi_{\text{old}}(a_t \mid s_t, g_t^{\text{emb}})}, 1-\epsilon, 1+\epsilon \right) \hat{A}_t 
		\Bigg) \Bigg],
	\end{aligned}
\end{equation}
where $\hat{A}_t$ is the estimated advantage function, $\epsilon$ is the clipping parameter, and $\mathcal{D}$ denotes the replay buffer or on-policy rollout distribution. The mask $\tilde{M}$ affects both the behavior policy $\pi_{\text{old}}$ and the updated policy $\pi_\theta$, ensuring consistency between sampling and optimization.

\section{Related Works}
\subsection{Open-World Environments}
Open-world environments~\citep{team2021open, cai2023open} are inherently challenging due to requirements for generalization, exploration, multi-objective optimization, and long-horizon planning and reasoning~\citep{hafner2022benchmarking, wang2023describe}.
There are three main approaches for applying RL in open-world environments in the existing literature. 
One approach is hierarchical reinforcement learning~\citep{hutsebaut2022hierarchical}, which simplifies complex decision-making processes by constructing a multi-level subtask structure.
However, due to the inherent limitations of reinforcement learning algorithms in planning and reasoning, the generalization and long-term decision-making capabilities of these methods are still constrained.
Another approach is model-based RL~\citep{moerland2023model, walker2023investigating}, which learns an explicit environment dynamics model to enable more sample efficient through simulated rollouts. 
However, these methods require learning an accurate world model, which results in significantly higher computational overhead, particularly in open-world environments with high-dimensional observations.
With the rapid development of LLMs, recent studies have explored integrating LLMs into RL pipelines~\citep{Zhou2024LLM4RL, schoepp2025evolving}. 
Leveraging their extensive prior knowledge, reasoning capabilities, and strong generalization, LLMs have been employed to provide high-level planning for RL agents~\citep{du2023guiding, zhang2024adarefiner, prakash2023llm, yan2025efficient}, which are discussed in detail in the following subsection.

\subsection{LLM-Enhanced RL}
LLM-enhanced RL~\citep{Zhou2024LLM4RL, schoepp2025evolving}, in which LLMs are employed as goal generators or policy selectors, with the core idea being to exploit their extensive prior knowledge for more effective task decomposition and decision-making.
To mention a few, SayCan~\citep{ichter2022do}, BOSS~\citep{zhang2023bootstrap} and When2Ask~\citep{hu2024enabling} utilize LLMs as skill planner to construct high-level plans or feasible skill sequences base on natural language instructions or task descriptions. 
However, these methods rely on assumed access to pretrained skills, and the generated plans lack structured representations, limiting their scalability and adaptability.
Furthermore, some works~\citep{pmlr-v202-hu23e, prakash2023llm, yan2025efficient} take a more direct straightforward approach, utilizing LLMs as policy teacher. 
Due to the fact that these methods either rely on a library of pretrained skills for high-level decision-making or depend on the LLMs' reasoning and language-to-action capabilities for low-level guidance, they typically require significant computational resources or extensive pretraining infrastructure. 
In addition, ELLM~\citep{du2023guiding}, AdaRefiner~\citep{zhang2024adarefiner}, LLMV-AgE~\citep{chi2025llmvage} and ~\citet{ruiz2024words} employ LLMs as goal generators to produce semantic subgoals that guide exploration. 
Unfortunately, due to their reliance on frequent and intensive LLM invocations, these methods suffer from low practical utility and poor computational efficiency.

\section{Experiments} 
In this section, experiments are conducted on two open-world RL benchmarks: Crafter~\citep{hafner2022benchmarking} and Craftax-Classic~\citep{matthews2024craftax}. The experiments aim to answer the following questions: 
1) How does the exploration efficiency of SGRL compare with existing LLM-enhanced RL methods?
2) How do goal-conditioned policy constraints contribute to the performance of SGRL?

To answer these questions, we compare SGRL against the following algorithms:
ELLM~\citep{du2023guiding}, which generates goals to guide agent exploration through online queries of an LLM; 
AdaRefiner~\citep{zhang2024adarefiner}, which enhances the quality of the LLM-generated goals by refining the prompts;
and PPO~\citep{schulman2017proximal}, which is a pure RL algorithm that does not involve an LLM.
Notably, for the specific hyperparameter settings of the PPO algorithm, we follow the \textit{Stable-Baselines3}~\citep{stable-baselines3}\footnote{available at \url{https://github.com/DLR-RM/stable-baselines3}} implementation for Crafter, and follow the official in \textit{Craftax} benchmark~\citep{matthews2024craftax}\footnote{available at \url{https://github.com/MichaelTMatthews/Craftax}} implementation for Craftax-Classic. 
In addition, human expert performance~\citep{hafner2022benchmarking} is included as a reference.

\subsection{Experimental Setup and Evaluation Metrics} 
\subsubsection{Environment}
Crafter is a widely used benchmark for open-world environments, evaluating agents on generalization, exploration, and long-term reasoning through 22 diverse achievements.
The Craftax-Classic environment re-implements Crafter in JAX.
Both benchmarks feature sparse rewards, complex goal hierarchies, and open-world exploration, making them ideal for evaluating our framework's ability to provide structured semantic guidance.
Further details of the environmental setup are provided in Appendix~\ref{Appendix A.1: Environments}.

The evaluation metrics from \cite{matthews2024craftax}, including success rate, score, return, and achievement depth, to comprehensively assess the performance of SGRL compared to the baseline algorithms. 
In addition, the training speed is reported in steps per second (SPS), presented in the tables as SPS ($\times 10^2$) for readability.
Further details can be found in Appendix~\ref{Appendix A.2: Evaluation Metrics}.

\subsubsection{Compute Resources}
Experiments on Crafter were conducted using a single A100 GPU with 40 GB of VRAM.
Experiments on Craftax-Classic were performed on a system equipped with an NVIDIA GeForce RTX 4090 (24 GB) and an Intel(R) Core(TM) i9-14900K CPU.
Results for both our algorithm and baseline methods are based on the same configurations.
All reported results are averaged over five random seeds and learning curves are smoothed over time.

\begin{figure*}[t]
	\centering
	\includegraphics[width=0.98\textwidth]{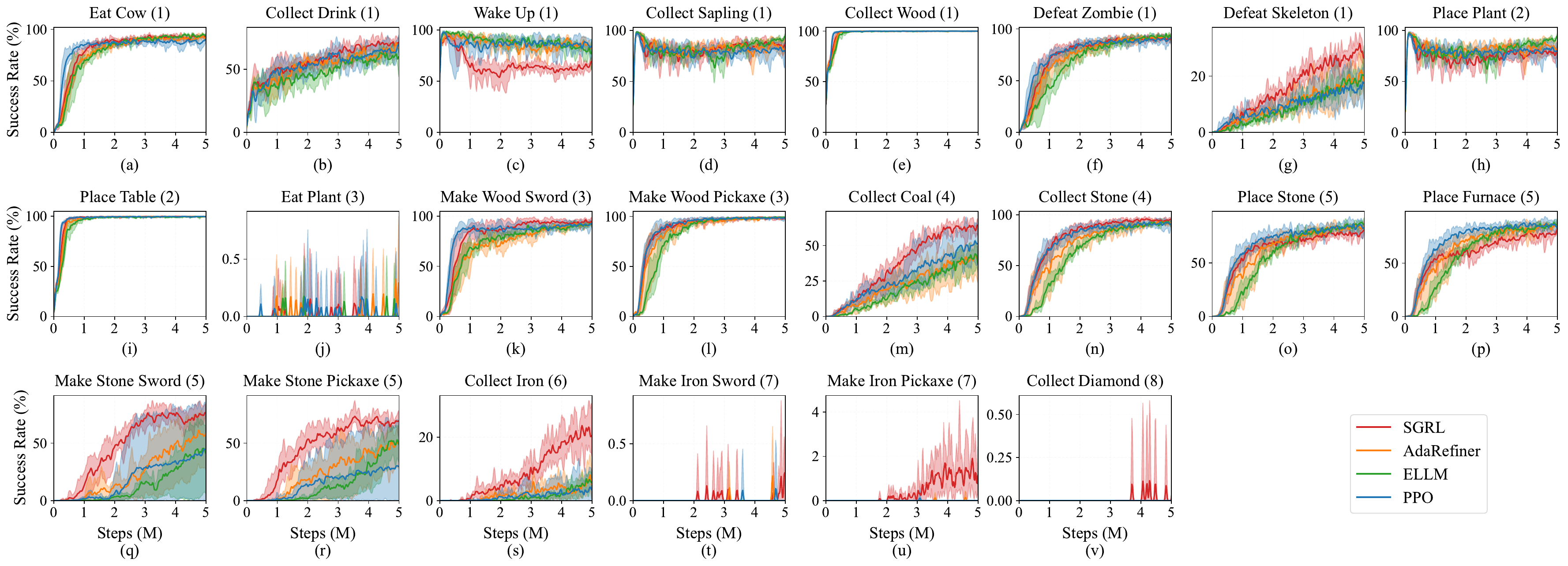} 
	\caption{Success rate curves for all achievements on Craftax-Classic. 
	Achievements are ranked based on their depth and the importance of unlocking them for subsequent tasks. 
	Achievements ranked later have greater depth and exert a stronger influence on subsequent achievements.
	A more intuitive version is shown in Figure \ref{AppendixFig: fig_22ach_tax_baseline_5M} in Appendix~\ref{Appendix D: Additional Main Results}.}
	\label{Fig: fig_tax_beseline_5M}
\end{figure*}

\begin{figure*}[t]
		\centering
		\includegraphics[width=0.98\textwidth]{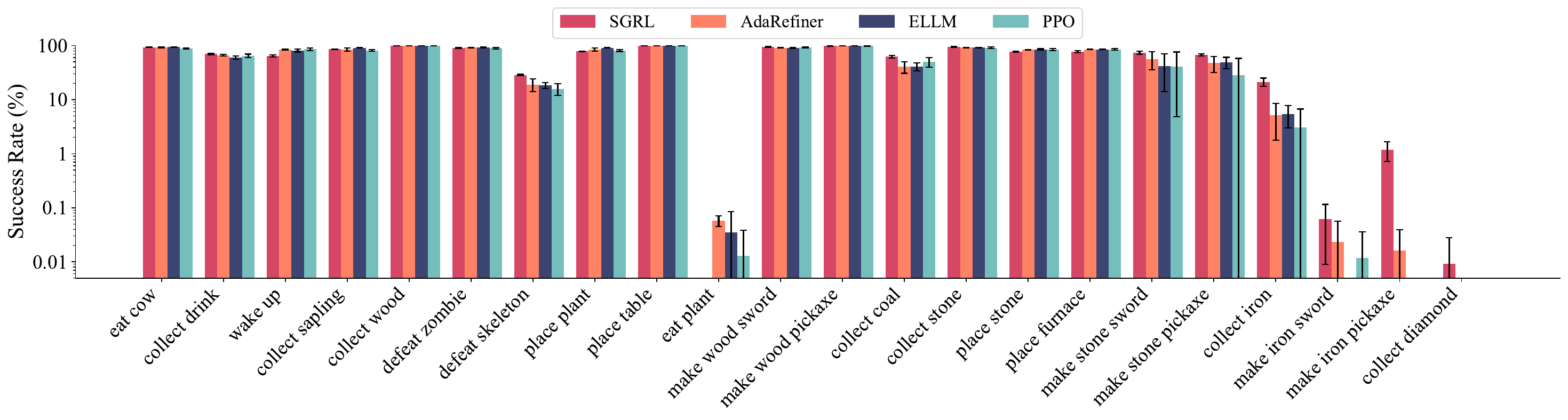} 
		\caption{Success rates across all achievements on Craftax-Classic at 5M steps.}
		\label{Fig: fig_bar_tax_beseline_5M}
\end{figure*}

\begin{table}[htb]
   \centering
   \setlength{\tabcolsep}{4pt}
   \begin{threeparttable}
        \label{Table: performance_comparison}
        \begin{tabular}{l c c c c}
            \toprule
			\textbf{Method} & \textbf{Score(\%)} & \textbf{Reward} & \makecell{\textbf{Achievement} \textbf{Depth}} &  \textbf{SPS ($\times 10^2$)} \\
            \midrule
            Human      & 50.5 $\pm$ 6.8   & 14.3 $\pm$ 2.3   & 8 & - \\
			\hline
            SGRL       & \textbf{33.8} $\pm$ 1.5   & \textbf{13.0} $\pm$ 0.3   & \textbf{8}  & 18.5 \\
            AdaRefiner &  28.5 $\pm$ 2.3  & 12.3 $\pm$ 0.9   & 7  & 0.3 \\
			ELLM       & 28.4 $\pm$ 2.5   & 12.2 $\pm$ 1.0   & 6  &  0.9\\
			PPO        & 24.8 $\pm$ 5.7 & 11.9 $\pm$ 1.1   & 6  & 135.3 \\
            \bottomrule
        \end{tabular}	
	\end{threeparttable}
	\caption{Performance of SGRL and baseline methods on Craftax-Classic at 5M steps.}
	\label{Tab: Craftax-Classic_perfor_5m}
\end{table}

\subsection{Main Results}\label{Subsection: Main Results}
Figure~\ref{Fig: fig_tax_beseline_5M} shows the success rate curves for all 22 achievements on Craftax-Classic.
From Figure~\ref{Fig: fig_tax_beseline_5M}, we can observe that SGRL consistently achieves higher success rates than the baselines when reaching the final few achievements (see Figure~\ref{Fig: fig_tax_beseline_5M} (q)-(v)).
However, for achievements like Wake Up (see Figure~\ref{Fig: fig_tax_beseline_5M} (c)) and Place Plant (see Figure~\ref{Fig: fig_tax_beseline_5M} (h)), which do not facilitate later exploration, SGRL's performance plateaus once high success rates are reached.
This phenomenon demonstrates that our goal-generation method produces farsighted objectives, enabling SGRL to transcend short-term rewards and maintain a stable and coherent policy in long-horizon decision-making tasks.
Moreover, as shown in Figure~\ref{Fig: fig_tax_beseline_5M} (v), SGRL successfully unlocks the \textit{Collect Diamond} achievement shortly after 3.7M steps, whereas AdaRefiner, ELLM, and PPO fail to achieve it even by 5M steps, demonstrating SGRL's effectiveness in enhancing exploration efficiency.
Figure~\ref{Fig: fig_bar_tax_beseline_5M} provides a more direct visualization, which clearly highlights SGRL's advantage in unlocking late-stage achievements, confirming its capability for effective long-horizon planning.
More experimental results can be found in Appendix~\ref{Appendix D: Additional Main Results}.

Table~\ref{Tab: Craftax-Classic_perfor_5m} summarizes the overall performance of SGRL and baseline methods on Craftax-Classic at 5M steps.
As demonstrated in Table~\ref{Tab: Craftax-Classic_perfor_5m}, SGRL outperforms AdaRefiner, ELLM, and PPO in terms of overall score and achievement depth on Craftax-Classic.
In addition, SGRL maintains a high reward level, yet does not exhibit a significant advantage in this regard. 
We attribute this to the misalignment between reward magnitude and exploration depth in the environment, which forces a trade-off between completing simple, reliably rewarded achievements and pursuing more challenging but potentially high-impact ones. 
This is evidenced by human performance: experts achieve very high scores without a corresponding increase in total reward. 
Further, SGRL requires only minimal LLM invocation, resulting in faster training speed.
Moreover, to better illustrate the performance of the proposed method, we provide results for various algorithms in the Crafter environment, which can be found in Appendix~\ref{Appendix D: Additional Main Results}.

\begin{figure*}[t]
	\centering
	\includegraphics[width=0.98\textwidth]{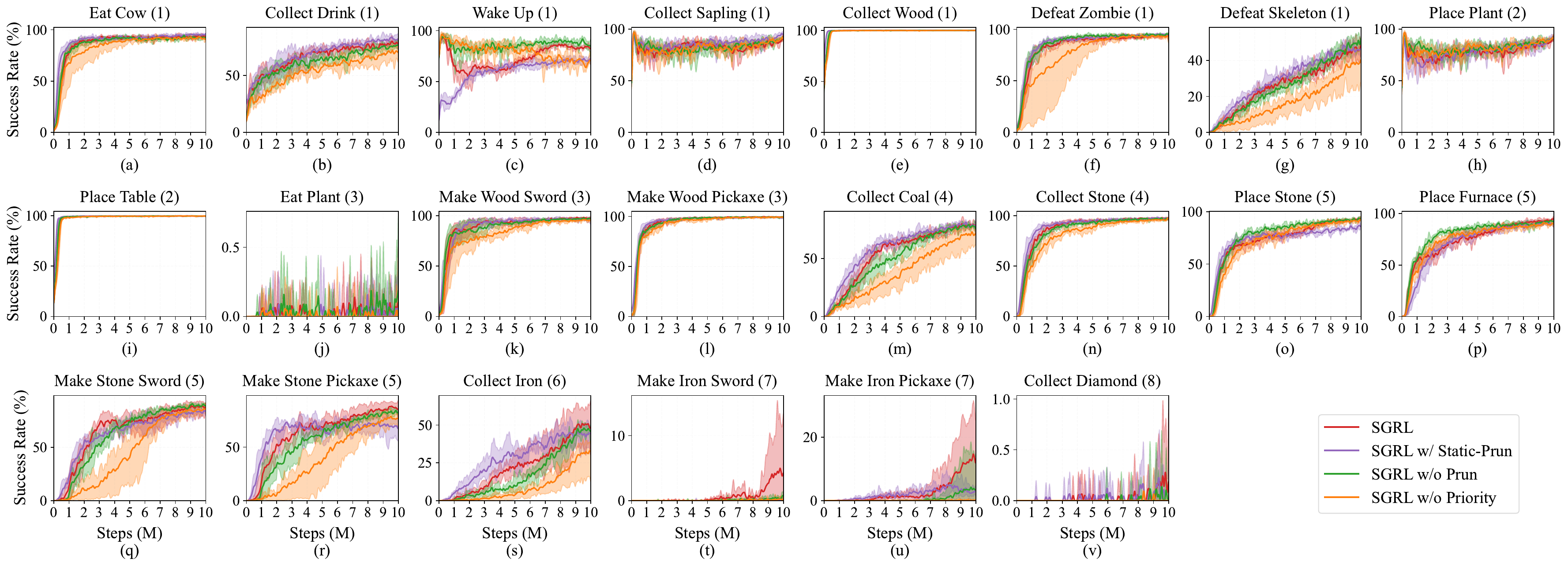} 
	\caption{Ablation Studies. Success rate curves for all achievements on Craftax-Classic.
	Achievements are ranked based on their depth and the importance of unlocking them for subsequent tasks. 
	Achievements ranked later have greater depth and exert a stronger influence on subsequent achievements.
	A more intuitive version is shown in Figure \ref{AppendixFig: fig_22ach_tax_abl_10M} in Appendix~\ref{Appendix E.1: Performance of Ablation Algorithm}.
	}
	\label{Fig: craftax_3*8_10m}
\end{figure*}

\begin{figure*}[t]
	\centering
	\includegraphics[width=0.98\textwidth]{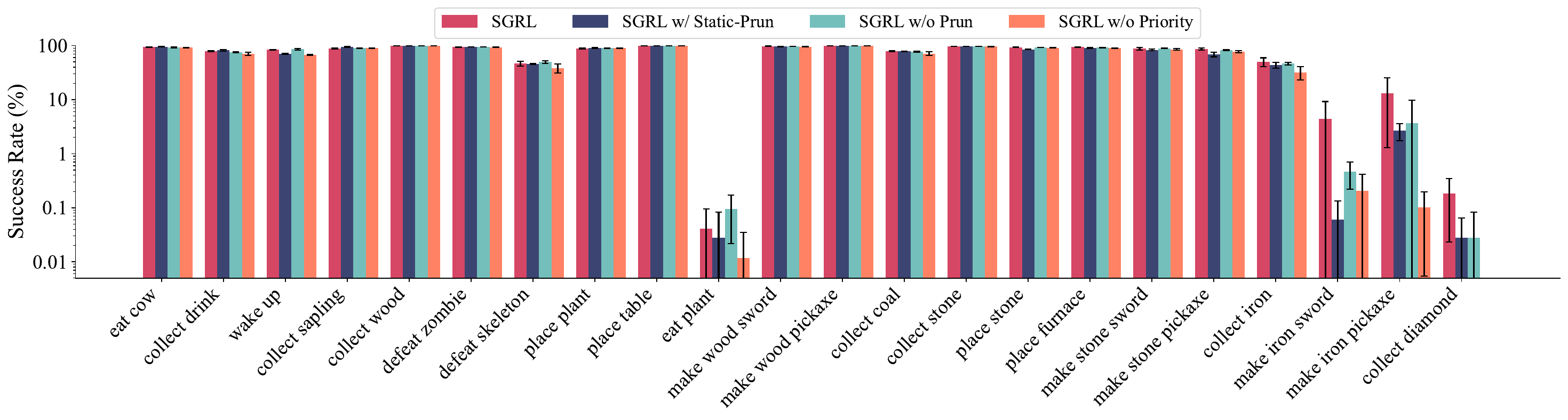} 
	\caption{Ablation Studies. Success rates across all achievements on Craftax-Classic at 10M steps.}
	\label{Fig: fig_bar_10M}
\end{figure*}

\subsection{Ablation Study}
\subsubsection{Ablation Variants}
To evaluate the contributions of each component of SGRL, we conducted ablation studies using three variants of SGRL on Craftax-Classic as follows:
(1) SGRL w/ Static-Prun: This variant retains the action pruning mechanism but replaces the adaptive pruning coefficient with a static masking scheme; 
(2)SGRL w/o Prun: This variant removes the goal-conditioned action pruner entirely;
(3)SGRL w/o Priority: This variant removes the priority assignment for goals.

\subsubsection{Ablation Analysis}
Figure~\ref{Fig: craftax_3*8_10m} shows the success rate curves of SGRL and its ablation variants for all 22 achievements on Crafter. 
From Figure~\ref{Fig: craftax_3*8_10m}, we can observe that SGRL significantly outperforms all variants on most of the deeper achievements after approximately 8M steps, indicating that both components of our algorithm contribute substantially to its performance.
Moreover, it is noteworthy that SGRL w/ Static-Prun, a variant of SGRL with strict action masking, exhibits strong early performance but is ultimately outperformed by SGRL in later stages, suggesting that over-reliance on the LLM may lead the agent to converge to suboptimal policies (see Figure~\ref{Fig: craftax_3*8_10m} (c), (o), (p) and (q)).

\begin{wraptable}{r}{0.65\textwidth}
	\vspace{-1em}
	\centering
	\small
	\setlength{\tabcolsep}{4pt}
	\begin{tabular}{lccc}
		\hline
		\textbf{Methd} & \textbf{Score (\%)}     & \textbf{Reward}         & \makecell{\textbf{Achievement} \\ \textbf{Depth}} \\
		\hline
		SGRL  & $\textbf{43.9} \pm 2.6$ & $\textbf{14.9} \pm 0.4$ & 8          \\
		SGRL w/ Static-Prun & $38.5 \pm 1.9 $         & $14.2 \pm 0.4 $         & 8           \\
		SGRL w/o Prun       & $40.0 \pm 2.1$          & $14.7 \pm 0.2$          & 8           \\
		SGRL w/o Priority   & $35.3 \pm 1.6$          & $13.9 \pm 0.6$          & 7          \\
		\hline
	\end{tabular}
	\caption{Ablation results on Craftax-Classic at 10M steps.}
	\label{Tab: craftax_10m}
\end{wraptable}

Figure~\ref{Fig: fig_bar_10M} and Table~\ref{Tab: craftax_10m} present the success rates, scores, rewards, and achievement depth across all 22 achievements in the Craftax-Classic environment at 10M steps.
The results demonstrate that SGRL achieves clear advantages on long-horizon tasks such as \textit{Make Iron Pickaxe} and \textit{Collect Diamond}, which require sequential planning and tool construction.
Notably, the design of goal prioritization plays a critical role in enabling SGRL to rapidly unlock deep achievements like \textit{Collect Diamond} by guiding the agent to focus on high-value, forward-looking goals during exploration.
More experimental results are in Appendix~\ref{Appendix E.1: Performance of Ablation Algorithm}. 

\begin{wrapfigure}{r}{0.65\textwidth}
	\vspace{-1em}
	\centering
	\includegraphics[width=0.6\textwidth]{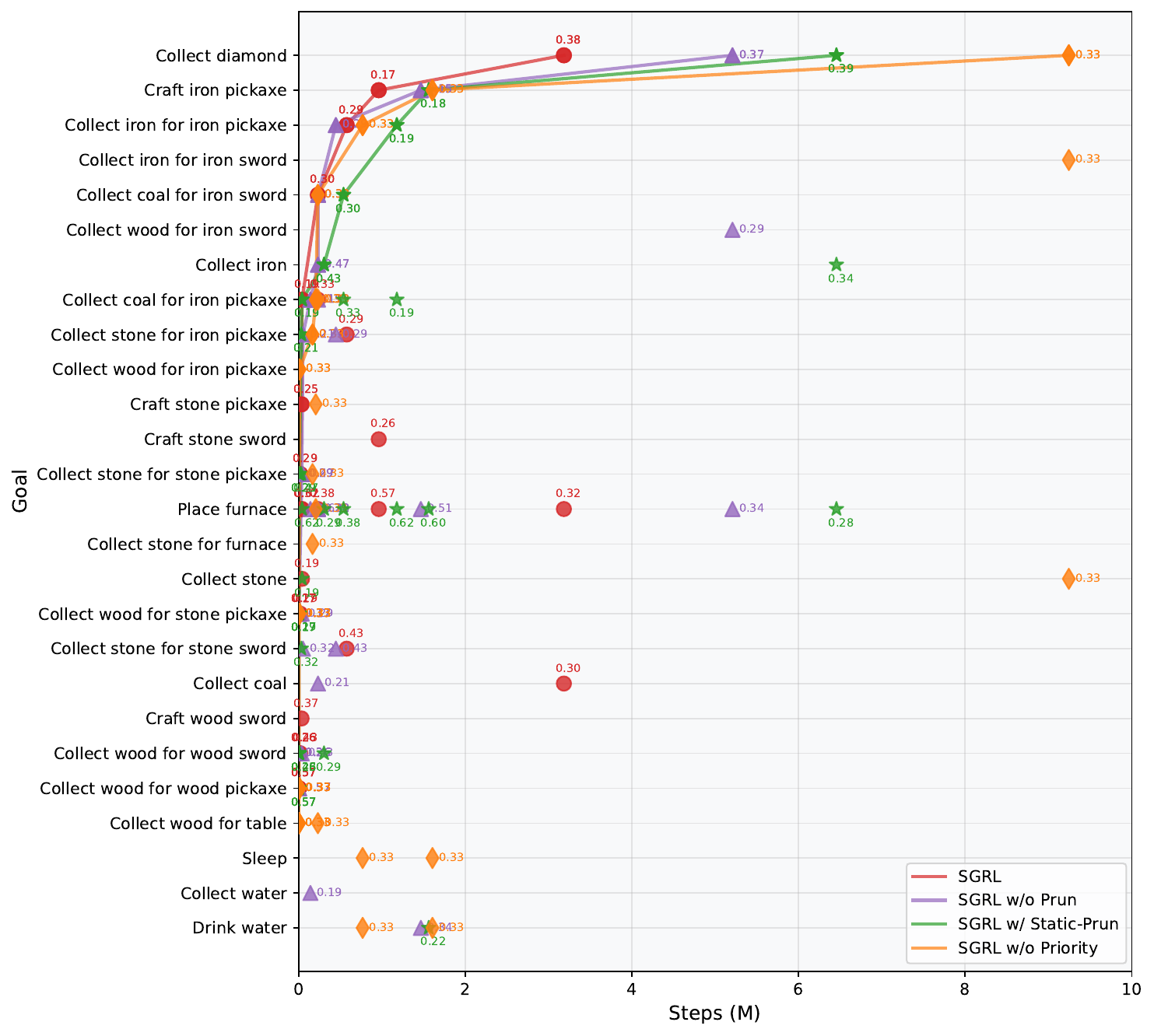} 
	\caption{Ablation Studies. Goals with priority weights on Craftax-Classic, where a higher-level goal is visualized only from the point it is assigned a non-zero weight.}
	\label{Fig: goal_prio}
	\vspace{-1em}
\end{wrapfigure}

Furthermore, for an in-depth analysis of SGRL's superiority, Figure~\ref{Fig: goal_prio} presents the goals with priority weights on Craftax-Classic. 
For clarity, a higher-level goal is visualized only from the point at which it is assigned a non-zero weight. 
The figure shows that SGRL consistently assigns a small priority weight to more forward-looking goals compared to the ablation algorithms at earlier stage, which motivates its agent to begin exploring more challenging achievements sooner. 
We hypothesize that this is the primary driving force behind SGRL's rapid attainment of superior performance.
For more detailed results, refer to Appendix~\ref{Appendix E.2: Heatmap of Goal with Priority Weights}, where we present the heatmap of goals with priority weights and provide a detailed analysis.

In summary, all ablation results collectively indicate that both the goal priority weights and action mask in our proposed SGRL play distinct roles.
Specifically, assigning priorities to goals within the structured goal planner is crucial for generating reasonable and effective goals. 
The goal-conditioned action pruner effectively enhances the agent's exploration capability in long-horizon tasks.

\section{Conclusion}
This paper proposes a novel LLM-enhanced RL method called SGRL that leverages LLMs to improve RL agents' exploration efficiency and long-horizon planning capabilities in open-world environments.
Specifically, we develop a structured goal planner that leverages the LLM to construct reusable goal-generating functions that select forward-looking goals and dynamically adjust their priority weights. 
Then, a goal-conditioned action pruner is designed to filter out actions misaligned with the goal, thereby guiding RL agents to select goal-consistent policies.
Finally, extensive experimental results demonstrate that SGRL achieves superior performance compared to existing LLM-enhanced RL baselines in terms of long-horizon planning and exploration efficiency.

\newpage


\bibliography{llm4rl}
\bibliographystyle{iclr2026_conference}

\newpage
\appendix
\section{Environments and Evaluation Metrics}  \label{Appendix A: Environments and Evaluation Metrics}
\subsection{Environments} \label{Appendix A.1: Environments}
\textit{Crafter~\citep{hafner2022benchmarking}} is a widely adopted benchmark for open-world RL, designed to assess agents' generalization, exploration, and long-term reasoning capabilities through 22 diverse achievements. 
These achievements are organized in a hierarchical dependency structure of up to 8 depth levels, where early-stage skills (e.g., collect wood, place table) unlock preconditions for increasingly complex tasks. 
The deepest and most challenging achievement (i.e., \textit{Collect Diamond}) requires agents to master long sequences of dependencies, from crafting stone tools to mining iron and finally accessing diamonds deep underground.
Crafter's procedurally generated environments further exacerbate challenges in sparse rewards, efficient exploration, and hierarchical planning, making it a strong testbed for evaluating structured goal-guided learning.

\textit{Craftax-Classic~\citep{matthews2024craftax}} is a high-performance, JAX-based reimplementation of Crafter that achieves a 250x simulation speedup via vectorization and parallelization. 
It faithfully reproduces Crafter's core task structure, environmental dynamics, and evaluation metrics, while enabling large-scale experimentation at 1B+ environment steps within practical compute budgets.

\subsection{Evaluation Metrics}  \label{Appendix A.2: Evaluation Metrics}
To demonstrate the effectiveness of our algorithm, we introduce the evaluation metrics as follows: 
\begin{itemize}
    \item \textit{Achievement Success Rate.} 
    This metric reflects the agent's learning capability and exploration depth by measuring the probability that each predefined achievement is successfully unlocked.
    \item \textit{Geometric Mean Score.}
    This metric reflects the balance between both easy and difficult goals.
    Following the official Crafter evaluation protocol~\citep{hafner2022benchmarking}, it is defined as:
    \begin{equation*}
        \text{score} = \exp\left(\frac{1}{N} \sum_{i=1}^{N} \log (1 + \varrho_i) \right) - 1,
    \end{equation*}
    where $\varrho_i \in [0,100]$ denotes the success rate of the $i$-th achievement, and $N$ is the total number of predefined achievements.
    \item \textit{Achievement Depth.}
    This metric measures the agent's exploration depth based on the furthest achievement it unlocks.
    \item \textit{Episode Return.}
    This metric reports the cumulative reward received per episode.
    \item \textit{Steps Per Second (SPS)}: This metric measures the number of environment steps processed per second, indicating the computational efficiency and speed of learning for each method.
\end{itemize}

\section{Implementation Details}  \label{Appendix B: Implementation Details}

\subsection{LLM}  \label{Appendix B.1: LLM}
We utilize \texttt{DeepSeek-R1}~\citep{guo2025deepseek} model to generate structured goal-generating planner code. 
Additionally, we use \texttt{DeepSeek-V3}~\citep{liu2024deepseek} model for selecting actions related to the defined goals. 
For all LLM queries, we follow the implementation of AdaRefiner~\citep{zhang2024adarefiner} to set the decoding parameters: a temperature of 0.5, top-p of 1.0, and a maximum token limit of 100. 
\texttt{DeepSeek-V3} model is employed to replicate the results of ELLM and AdaRefiner.

\subsection{Text Embedding} \label{Appendix B.2: Text Embedding}
For text embedding, we use \texttt{paraphrase-MiniLM-L6-v2}~\citep{reimers2019sentence} model as the encoder.

\subsection{Prompt Design}  \label{Appendix B.3: Prompt Dsign}

\subsubsection{Prompt Design for Structured Goal Planner}
In the Structured Goal Planner, the large model is employed to generate goal-generation code and to update the priorities of goals. To enable the LLM to produce high-quality code, we adopt a multi-stage prompting process:
\begin{itemize}
    \item \textbf{Design Stage.}
    At this stage, the model is asked to first design the class structure and key functional modules according to the task requirements.
    \item \textbf{Implementation Stage.}
    After the design, the model is prompted to output detailed, complete Python code that follows PEP8 standards with clear logic and sufficient comments.
    \item \textbf{Reflection and Revision Stage.}
    Finally, the model is prompted to reflect on the generated code, identify potential issues, and provide corrections or optimizations.
\end{itemize}
To guide the model in updating goal priorities, the prompt additionally specifies that every 2 million steps the LLM should update the priority values of goals within the generated code. This ensures that goal selection remains dynamic and aligned with the agent's current objectives.

\begin{tcolorbox}[colback=white, colframe=black, sharp corners, boxrule=1pt, title=Prompt Template for Structured Goal Planner Design, breakable,
    enhanced, title after break={Prompt Template for Structured Goal Planner Design (continued)}]
	Crafter is a 2D open-world survival game with visual input; its world is procedurally generated. Players must search for food and water, find shelter to sleep, defend against monsters, gather materials, and craft tools. Crafter's objective is to evaluate an agent's capabilities through a series of semantically meaningful achievements that can be unlocked in each playthrough---for example, discovering resources and crafting tools. Consistently unlocking all achievements requires strong generalization, deep exploration, and long-term reasoning.

	You are an experienced Python developer. The task is to create a key functional module of an advanced goal-generation system that can dynamically produce prioritized goals based on textual environment observations. The system must include functions for survival-need assessment, a tool-crafting tree, resource-collection configuration, and achievement tracking.

	You should primarily design the \texttt{OptimizedGoalGenerator} class structure and its key functional modules. The Agent will call the \texttt{determine\_goal} function to obtain goals:
\begin{lstlisting}[
	language=Python,
	breaklines=true,
	basicstyle=\footnotesize\ttfamily,  
	keywordstyle=\bfseries\color{blue},
	commentstyle=\color{gray},
	stringstyle=\color{red},
	showstringspaces=false,
	xleftmargin=0em,                   
	frame=single,                     
	backgroundcolor=\color{gray!5},    
	breakatwhitespace=true            
]
def determine_goal(self, text_obs):
    return top_three_goal
\end{lstlisting}

	Each \texttt{goal} is a dictionary of the form:
\begin{lstlisting}[
	language=Python,
	breaklines=true,
	basicstyle=\footnotesize\ttfamily,  
	keywordstyle=\bfseries\color{blue},
	commentstyle=\color{gray},
	stringstyle=\color{red},
	showstringspaces=false,
	xleftmargin=0em,                   
	frame=single,                     
	backgroundcolor=\color{gray!5},    
	breakatwhitespace=true            
]
{'goal':    ,
 'priority':    , }
\end{lstlisting}

	The state of environmental text is represented by \texttt{text\_obs}: \\
	\textbf{Example 1:}\\
	You see: plant, zombie, tree, grass, sand, path, stone\\
	Inventory: wood: 1\\
	Status: health: 11\%, Fullness: 0\%, Hydration: 0\%, Wakefulness: 88\%\\
	Sky brightness level: 68\%
	\textbf{Example 2:}\\
	You see: plant, tree, grass, path, stone\\
	Inventory: \\
	Status: health: 99\%, Fullness: 77\%, Hydration: 66\%, Wakefulness: 77\%\\
	Sky brightness level: 99\%\\

	Now please provide the \texttt{OptimizedGoalGenerator} class structure and its key functional modules.

\begin{lstlisting}[
	language=Python,
	breaklines=true,
	basicstyle=\footnotesize\ttfamily,  
	keywordstyle=\bfseries\color{blue},
	commentstyle=\color{gray},
	stringstyle=\color{red},
	showstringspaces=false,
	xleftmargin=0em,                   
	frame=single,                     
	backgroundcolor=\color{gray!5},    
	breakatwhitespace=true            
]
class OptimizedGoalGenerator:
    def __init__(self):
        ...
    def determine_goal(self, text_obs):
        ...
\end{lstlisting}

\end{tcolorbox}

\begin{tcolorbox}[colback=white, colframe=black, sharp corners, boxrule=1pt, title=Prompt Template for Structured Goal Planner Implementation, breakable,
    enhanced, title after break={Prompt Template for Structured Goal Planner Implementation (continued)}]

Crafter is a 2D open-world survival game with visual input, where the world is procedurally generated. Players need to find food and water, locate a place to sleep, defend against monsters, gather materials, and craft tools. The objective of Crafter is to evaluate an agent's capabilities through a series of semantically meaningful achievements, which can be unlocked in each game session, such as discovering resources and crafting tools. Continuously unlocking all achievements requires strong generalization, deep exploration, and long-term reasoning.

As a Python expert, your task is to create an advanced goal generation system that dynamically generates prioritized goals based on environmental observation text. The system should include survival needs assessment, a crafting dependency tree, resource collection configuration, and achievement tracking.

Environment Details:

\textbf{Items:} sapling, wood, stone, coal, iron, diamond, wood\_pickaxe, stone\_pickaxe, iron\_pickaxe, wood\_sword, stone\_sword, iron\_sword (all with max: 9, initial: 0)

\textbf{Collectable resources:} tree, stone, coal, iron, diamond, water, grass (with required tools, output, and leaves defined)

\textbf{Placable objects:} stone, table, furnace, plant (with usage, location, and type defined)

\textbf{Craftable tools:} wood\_pickaxe, stone\_pickaxe, iron\_pickaxe, wood\_sword, stone\_sword, iron\_sword (with required materials, nearby crafting stations, and output quantity)

\textbf{Achievements:} collect\_coal, collect\_diamond, collect\_drink, collect\_iron, collect\_sapling, collect\_stone, collect\_wood, defeat\_skeleton, defeat\_zombie, eat\_cow, eat\_plant, make\_iron\_pickaxe, make\_iron\_sword, make\_stone\_pickaxe, make\_stone\_sword, make\_wood\_pickaxe, make\_wood\_sword, place\_furnace, place\_plant, place\_stone, place\_table, wake\_up

Environment Text Rendering Function:

\begin{lstlisting}[
	language=Python,
	breaklines=true,
	basicstyle=\footnotesize\ttfamily,  
	keywordstyle=\bfseries\color{blue},
	commentstyle=\color{gray},
	stringstyle=\color{red},
	showstringspaces=false,
	xleftmargin=0em,                   
	frame=single,                     
	backgroundcolor=\color{gray!5},    
	breakatwhitespace=true            
]
def render_craftax_text_describ_2(self, view_arr, index):
    (map_view, mob_map, inventory_values, status_values) = view_arr

    mob_id_to_name = ["zombie", "cows", "skeletons", "arrows"]
    block_id_to_name = ["invalid", "out of bounds", "grass", "water", "stone", "tree", "wood", "path",
                        "coal", "iron", "diamond", "crafting table", "furnace", "sand", "lava", "plant", "ripe plant"]

    text_view_values = set()

    block_names = np.vectorize(lambda x: block_id_to_name[x])(map_view[index])
    text_view_values.update(block_names.flatten())

    mob_ids = np.argmax(mob_map[index], axis=-1)
    mob_names = np.vectorize(lambda x: mob_id_to_name[x])(mob_ids)
    mob_mask = mob_map[index].max(axis=-1) > 0.5
    text_view_values.update(mob_names[mob_mask].flatten())
    text_view = ", ".join(text_view_values)

    inv_names = ["wood", "stone", "coal", "iron", "diamond", "sapling",
                 "wood pickaxe", "stone pickaxe", "iron pickaxe",
                 "wood sword", "stone sword", "iron sword"]
    text_obs_inv = ", ".join([f"{name}: {value}"
                              for name, value in zip(inv_names, inventory_values[index])
                              if value > 0])

    status_names = ["Health", "Fullness", "Hydration", "Wakefulness", "Sky brightness level"]
    status = ", ".join([f"{name}: {int(value / 0.09)}%"
                        for name, value in zip(status_names, status_values[index])])

    text_obs = "You see: " + text_view + "\nInventory: " + text_obs_inv + "\nStatus: " + status
    return text_obs
\end{lstlisting}

Example Environmental Text Observations:

    Example 1:\\
    You see: plant, zombie, tree, grass, sand, path, stone\\
    Inventory: wood: 1\\
    Status: health: 11\%, Fullness: 0\%, Hydration: 0\%, Wakefulness: 88\%, Sky brightness level: 68\%
    
    Example 2:\\
    You see: plant, tree, grass, path, stone\\
    Inventory: \\
    Status: health: 99\%, Fullness: 77\%, Hydration: 66\%, Wakefulness: 77\%, Sky brightness level: 99\%

You have already designed the code architecture. Your goal is to complete this code and create an advanced goal generation system capable of dynamically generating prioritized goals based on environmental observation text.

The code architecture: \\
\{last\_llm\_response\}

\end{tcolorbox}

\begin{tcolorbox}[colback=white, colframe=black, sharp corners, boxrule=1pt, title=Prompt Template for Structured Goal Planner Reflection and Revision, breakable,
    enhanced, title after break={Prompt Template for Structured Goal Planner Reflection and Revision (continued)}]

You are an expert Python developer and code reviewer for Crafter's goal generation system. Your task is to critically analyze the previously designed goal planner and provide an evaluation. For each submitted code version:

\begin{itemize}
    \item If the code is complete, correct, and efficiently implements all required functionalities (survival needs assessment, resource collection, crafting, achievement tracking, threat handling, and goal prioritization), label it as \textbf{good}.
    \item If there are issues, missing features, or opportunities for optimization, label it as \textbf{bad}, and provide a clear explanation of the problems.
\end{itemize}

After evaluation, if the code is labeled \textbf{bad}, generate a fully optimized and corrected version of the code that addresses all identified issues. The optimized code should:

\begin{itemize}
    \item Correctly handle all environmental observations and edge cases.
    \item Properly assess and prioritize survival needs.
    \item Integrate resource collection, crafting, and achievement goals with correct dependencies.
    \item Handle threats and defensive behaviors appropriately.
    \item Be modular, readable, and maintainable, following Python best practices.
\end{itemize}

Please review the following Structured Goal Planner code. Evaluate its quality: provide the label \textbf{good} if it is fully correct and functional, or \textbf{bad} if improvements are needed. For \textbf{bad} code, explain the deficiencies clearly and provide a complete, optimized version of the code that fixes all issues while preserving the intended functionality.

The goal-generated code:
\{{last\_llm\_response}\}

\end{tcolorbox}

\begin{tcolorbox}[colback=white, colframe=black, sharp corners, boxrule=1pt, title=Prompt Template for Updating Goal Priority Values, breakable,
    enhanced, title after break={Prompt Template for Updating Goal Priority Values (continued)}]

You are the Goal Priority Analyst for Crafter games. Your task is to update the priority weights of goals in the goal generation code, based on the current state and action trajectory of the intelligent agent. You act as a specialized assistant whose only responsibility is to adjust priority values to improve goal selection; do not change any code logic or structure.

\textbf{Inputs provided to you:}
\begin{itemize}
    \item \textbf{Goal generation code}: A Python code module that defines goals, their attributes, and initial priority weights. (\{goal\_code.py\})
    \item \textbf{Agent state and trajectory}: A structured representation of the current state of the intelligent agent, including completed goals, actions taken, and environment status. (\{agent\_state.json\})
\end{itemize}

\textbf{Your instructions:}
\begin{itemize}
    \item Read the goal generation code and the agent state/trajectory.
    \item Update the numeric \textbf{priority weights} in the code to reflect the current importance of each goal.
    \item Do \textbf{not} change any function definitions, logic, or class structures. Only modify numeric values associated with goal priorities.
    \item Ensure the updated code is fully executable and maintains its original structure.
    \item Keep all interfaces unchanged so that the planner can call the updated code directly.
\end{itemize}

\textbf{Example Workflow:}
\begin{enumerate}
    \item Load the goal generation code and parse the goals.
    \item Analyze the agent's current state and past actions.
    \item Determine new priority weights for each goal.
    \item Replace only the priority numbers in the code with the updated values.
    \item Output the complete updated Python code.
\end{enumerate}

\textbf{Output Format:}
\begin{itemize}
    \item Return the entire Python code as a single code block.
    \item Ensure all class and function definitions remain intact.
    \item Only the numeric priority values are changed.
\end{itemize}

\end{tcolorbox}

\subsubsection{Prompt Design for Goal-Conditioned Action Pruner}

In the Goal-Conditioned Action Pruner, the large model selects actions that are directly relevant to a given goal through the use of prompts, thereby enabling goal-driven behavior. To guide the model in generating goal-consistent actions, the prompt explicitly defines the meaning of each action in the action space and provides examples illustrating which actions should be chosen for specific goals. In this way, the model can effectively filter actions that align with the goal, ensuring that the agent maintains coherent and goal-directed behavior during execution.

\begin{tcolorbox}[colback=white, colframe=black, sharp corners, boxrule=1pt, title=Prompt Template for Goal-Conditioned Action Pruner Implementation, breakable,
    enhanced, title after break={Prompt Template for Goal-Conditioned Action Pruner Implementation (continued)}]

You are the goal analyst for Crafter games, and a goal planner provides goal guidance for game characters. The agent needs to perform one or more steps to achieve this goal, and you help the agent choose the appropriate actions to accomplish it.

\textbf{Tips:}
\begin{itemize}
    \item The goal is something that the intelligent agent is currently capable of executing under certain conditions.
    \item The intelligent agent may need to move to a certain location to trigger the execution condition.
    \item The action space consists of the following 17 actions:
    \begin{enumerate}
        \item noop \# do nothing
        \item move\_left
        \item move\_right
        \item move\_up
        \item move\_down
        \item do
        \item sleep
        \item place\_stone
        \item place\_table
        \item place\_furnace
        \item place\_plant
        \item make\_wood\_pickaxe
        \item make\_stone\_pickaxe
        \item make\_iron\_pickaxe
        \item make\_wood\_sword
        \item make\_stone\_sword
        \item make\_iron\_sword
    \end{enumerate}
    \item Among them, the action \texttt{noop} means do nothing.
    \item The action \texttt{do} means it can complete the following: eat plant, attack zombie, attack skeleton, attack cow, chop tree for wood, mine stone, mine coal, mine iron, mine diamond, drink water, chop grass for sapling.
\end{itemize}

\textbf{Examples:}
\begin{itemize}
    \item Goal: \{Mine Iron\} \\
    Related actions: \{move\_left, move\_right, move\_up, move\_down, do\}
    \item Goal: \{make stone pickaxe\} \\
    Related actions: \{move\_left, move\_right, move\_up, move\_down, make\_stone\_pickaxe\}
    \item Goal: \{sleep\} \\
    Related actions: \{sleep\}
    \item Goal: \{attack cow\} \\
    Related actions: \{move\_left, move\_right, move\_up, move\_down, do\}
\end{itemize}

For each given goal, generate a set of feasible actions from the action space. Include any movements or execution actions that can reasonably help achieve the goal. Focus on feasibility rather than strict optimality.

The current goal is: {goal}.

Please select actions that are relevant to the goal.

\end{tcolorbox}

\begin{table}[h]
	\centering
	\begin{tabular}{lcc}
		\hline
		\textbf{Parameter}         & \textbf{Value (Crafter)} & \textbf{Value (Craftax-Classic)}\\
		\hline
		Training Steps             & \{1M, 5M\}     & \{5M, 10M\}      \\
		Learning Rate              & 7e-4           & 7e-4             \\
		Optimizer                  & Adam           & AdamW            \\
		Batch Size                 & 128            & 512              \\
		Number of Envs             & 1              & 256              \\
		Update Epochs              & 16             & 4                \\
		Clip Ratio                 & 0.2            & 0.2              \\
		Discount Factor $\gamma$   & 0.97           & 0.97             \\
		Entropy Coefficient        & 0.01           & 0.01             \\
		Value Function Coefficient & 0.5            & 0.5              \\
		Activation Function        & ReLU           & ReLU             \\
		\hline
	\end{tabular}
	\caption{Hyperparameters for PPO}
	\label{Tab: ppo_hyper}
\end{table}

\subsection{PPO Algorithm}  \label{Appendix B.4: PPO Algorithm} 
For the specific hyperparameter settings of the PPO algorithm, we follow the \textit{Stable-Baselines3}~\citep{stable-baselines3}\footnote{Available at \url{https://github.com/DLR-RM/stable-baselines3}} implementation for Crafter, and follow the official in \textit{Craftax} benchmark~\citep{matthews2024craftax}\footnote{Available at \url{https://github.com/MichaelTMatthews/Craftax}} implementation for Craftax-Classic. 
Since SGRL, ELLM, and AdaRefiner are all implemented based on PPO, we use the same core PPO hyperparameters, as shown in Table~\ref{Tab: ppo_hyper}.

\section{Additional Preliminary Results}  \label{Appendix C: Additional Preliminary Results} 
\begin{figure}[t]
	\centering
	\includegraphics[width=0.98\textwidth]{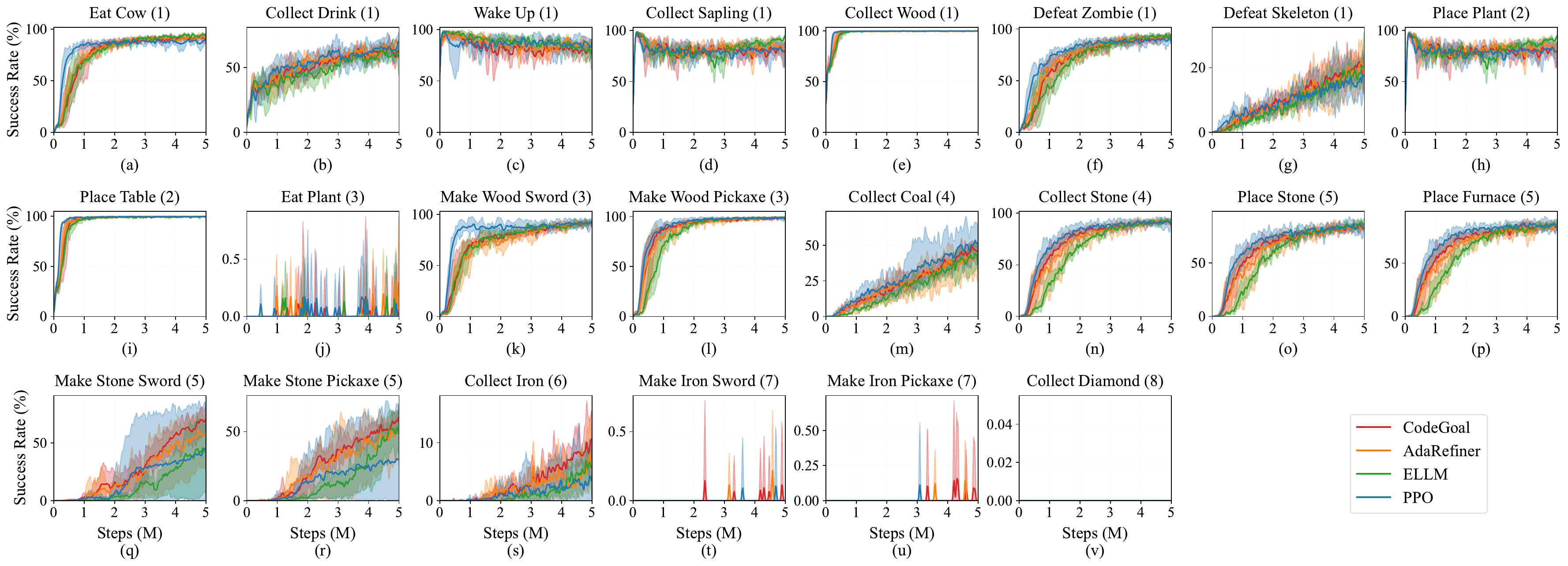} 
	\caption{Preliminary Results.
	Success rate curves for all achievements on Craftax-Classic within 5M steps.
	We rank the achievements based on their depth and the importance of unlocking them for subsequent tasks. 
	Achievements ranked later have greater depth and exert a stronger influence on subsequent achievements.
	A more intuitive version is shown in Figure \ref{AppendixFig: fig_22ach_tax_mot_5M}.
    }
	\label{AppendixFig: fig_tax_mot_5M}
\end{figure}

\begin{figure}[t]
	\centering
	\includegraphics[width=0.98\textwidth]{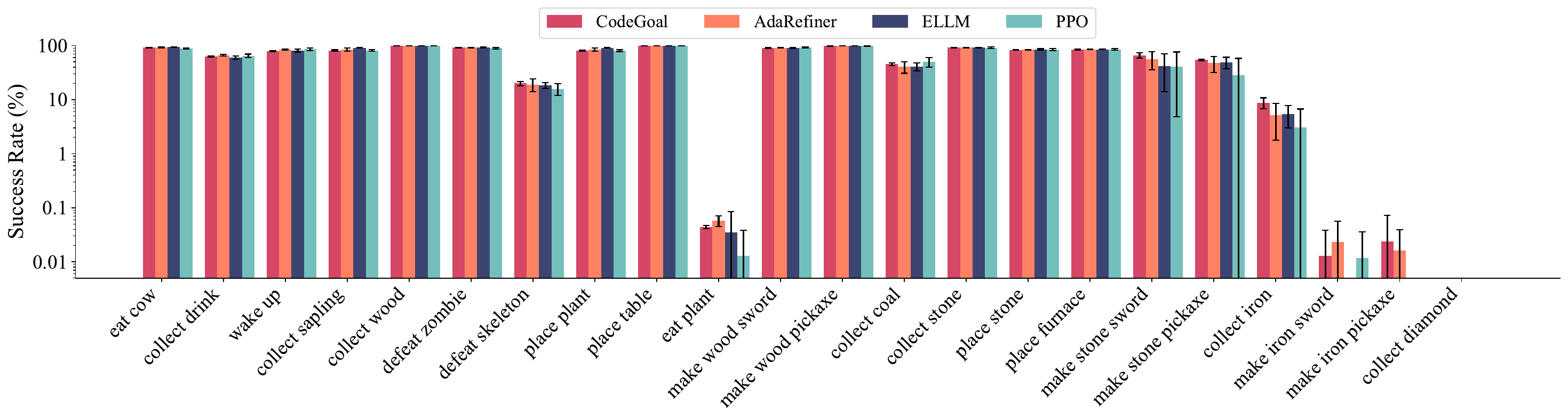} 
	\caption{Preliminary Results.
	Success rates across all Craftax-Classic achievements at 5M steps.}
	\label{AppendixFig: fig_tax_mot_bar_5M}
\end{figure}

\begin{figure}[t]
	\centering
	\includegraphics[width=0.98\textwidth]{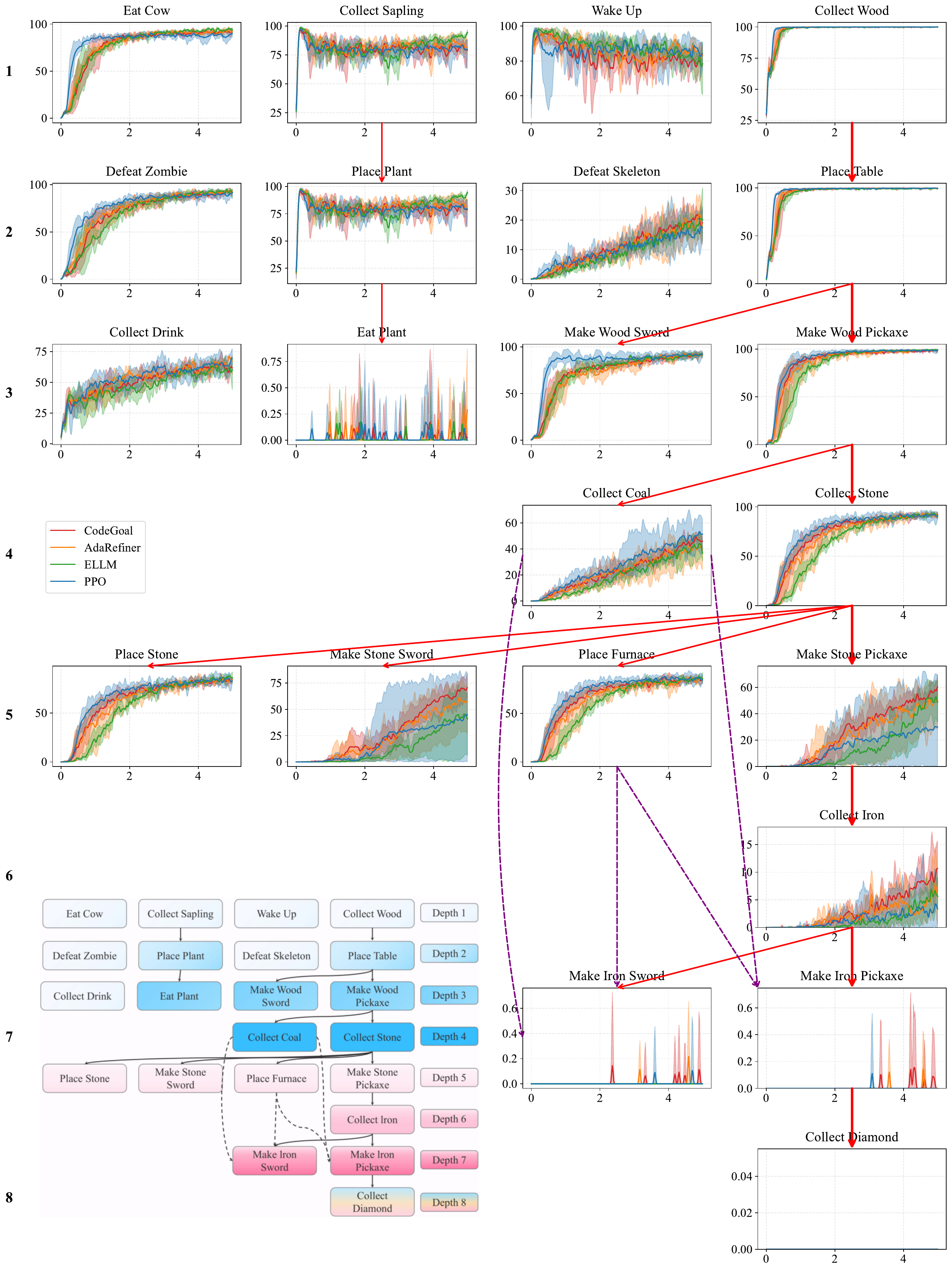}
	\caption{Preliminary Study.
	Success rate curves for all achievements on Craftax-Classic within 5M steps.
	Solid and dashed arrows indicate direct and cross-depth dependencies, respectively. 
	The bottom-left panel visualizes the full achievement dependency graph, with achievement depth encoded by color (depth 1-8 from top to bottom). 
	}
	\label{AppendixFig: fig_22ach_tax_mot_5M}
	\vspace{-1em}
\end{figure}

Figures \ref{AppendixFig: fig_tax_mot_5M}-\ref{AppendixFig: fig_22ach_tax_mot_5M} present more detailed results of the preliminary study experiments on Craftax-Classic.

\section{Additional Main Results}   \label{Appendix D: Additional Main Results}
\begin{figure}[h]
	\centering
	\begin{subfigure}[b]{0.98\textwidth}
		\centering
		\includegraphics[width=\textwidth]{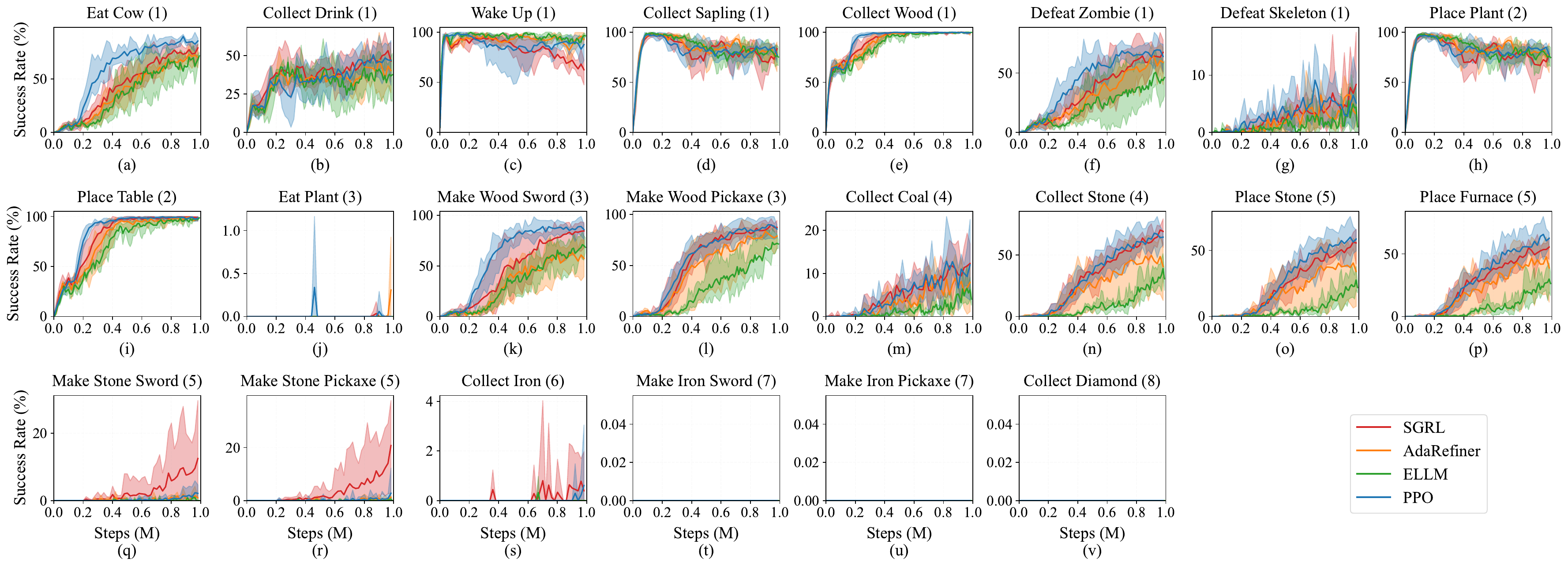} 
		\caption{1M Steps}
		\label{AppendixFig: fig_tax_beseline_1M}
	\end{subfigure}

	\begin{subfigure}[b]{0.98\textwidth}
		\centering
		\includegraphics[width=\textwidth]{fig_tax_beseline_5M.pdf} 
		\caption{5M Steps}
		\label{AppendixFig: fig_tax_beseline_5M}
	\end{subfigure}

	\begin{subfigure}[b]{0.98\textwidth}
		\centering
		\includegraphics[width=\textwidth]{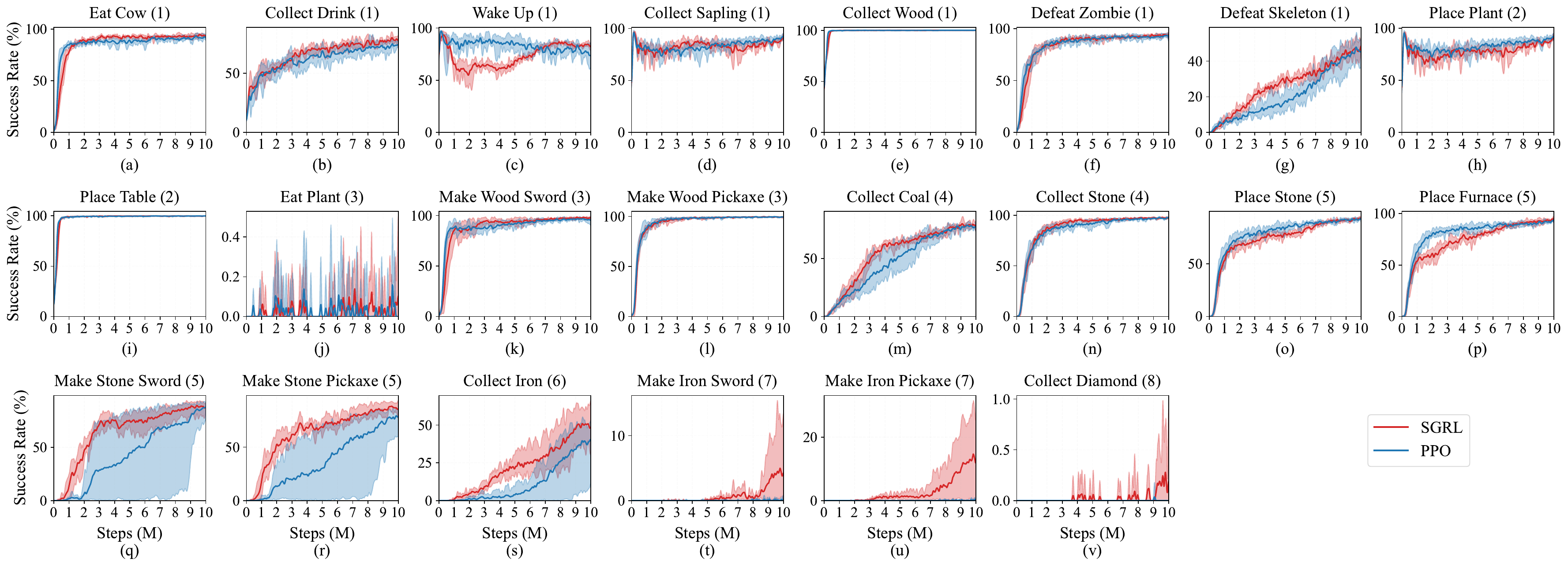} 
		\caption{10M Steps}
		\label{AppendixFig: fig_tax_beseline_10M}
	\end{subfigure}
	\caption{Main Results.
	Success rate curves for all achievements on Craftax-Classic within different training steps.
	We rank the achievements based on their depth and the importance of unlocking them for subsequent tasks. 
	Achievements ranked later have greater depth and exert a stronger influence on subsequent achievements.
	A more intuitive version is shown in Figure \ref{AppendixFig: fig_22ach_tax_baseline_5M}.
    }
	\label{AppendixFig: fig_tax_beseline}
\end{figure}

\begin{figure}[h]
	\vspace{3em}
	\begin{subfigure}[b]{0.98\textwidth}
		\centering
		\includegraphics[width=\textwidth]{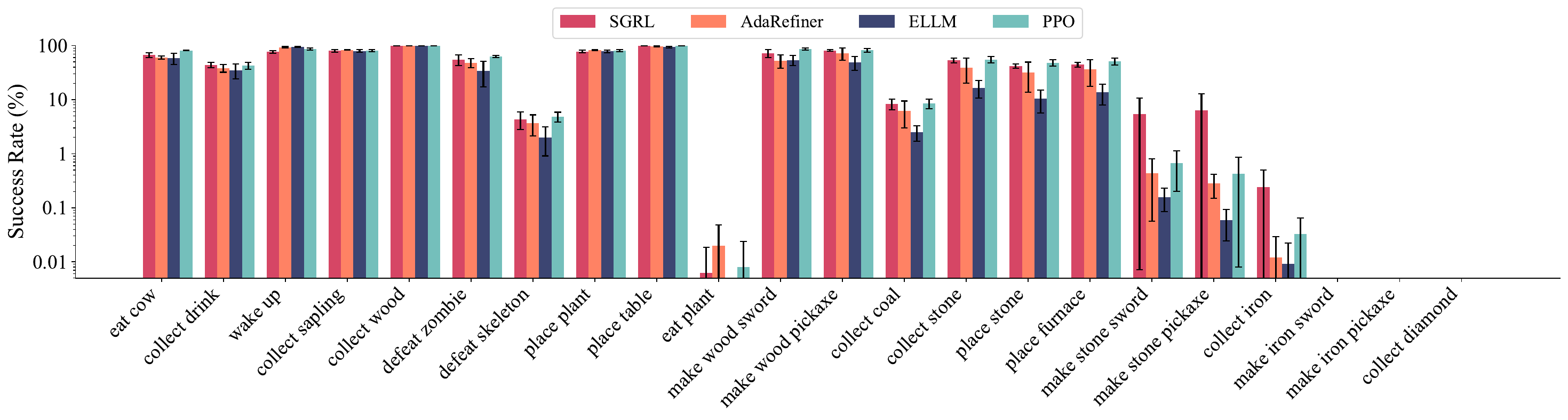} 
		\caption{1M Steps}
		\label{AppendixFig: fig_bar_tax_beseline_1M}
	\end{subfigure}

	\begin{subfigure}[b]{0.98\textwidth}
		\centering
		\includegraphics[width=\textwidth]{fig_bar_tax_beseline_5M.pdf} 
		\caption{5M Steps}
		\label{AppendixFig: fig_bar_tax_beseline_5M}
	\end{subfigure}

	\begin{subfigure}[b]{0.98\textwidth}
		\centering
		\includegraphics[width=\textwidth]{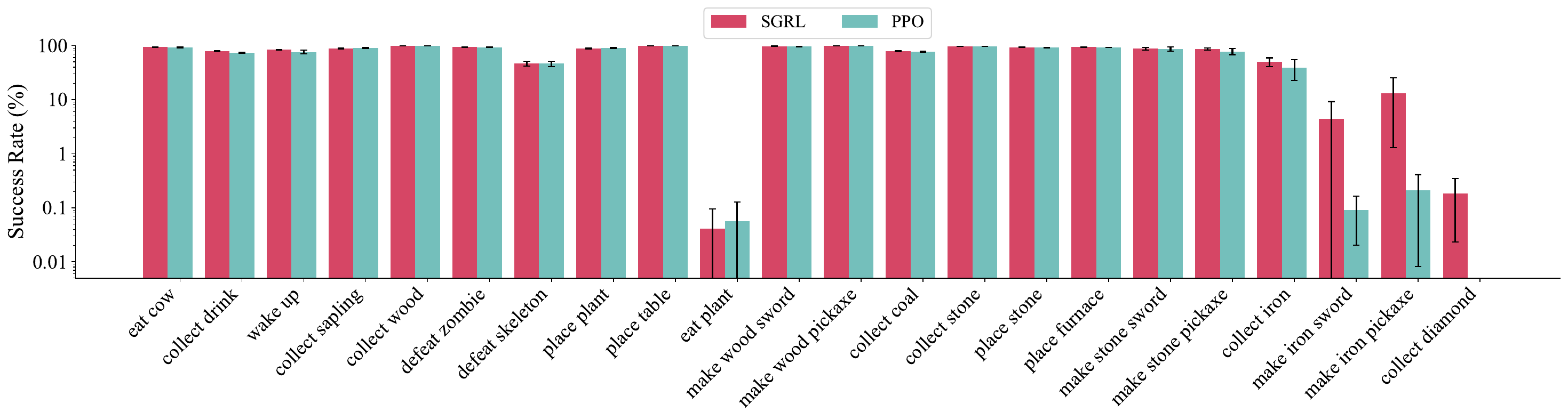} 
		\caption{10M Steps}
		\label{AppendixFig: fig_bar_tax_beseline_10M}
	\end{subfigure}
	\caption{Main Results.
	Success rates across all achievements on Craftax-Classic at different training steps.
	\textbf{Note:} Since ELLM and AdaRefiner require frequent online calls to the LLM (\texttt{DeepSeek-V3}) during training, they incur substantial computational costs and training time. 
	Therefore, we only reproduce the results within 5M steps.
	}
	\label{AppendixFig: fig_bar_tax_beseline}
\end{figure}

\begin{figure}[h]
	\centering
	\includegraphics[width=0.98\textwidth]{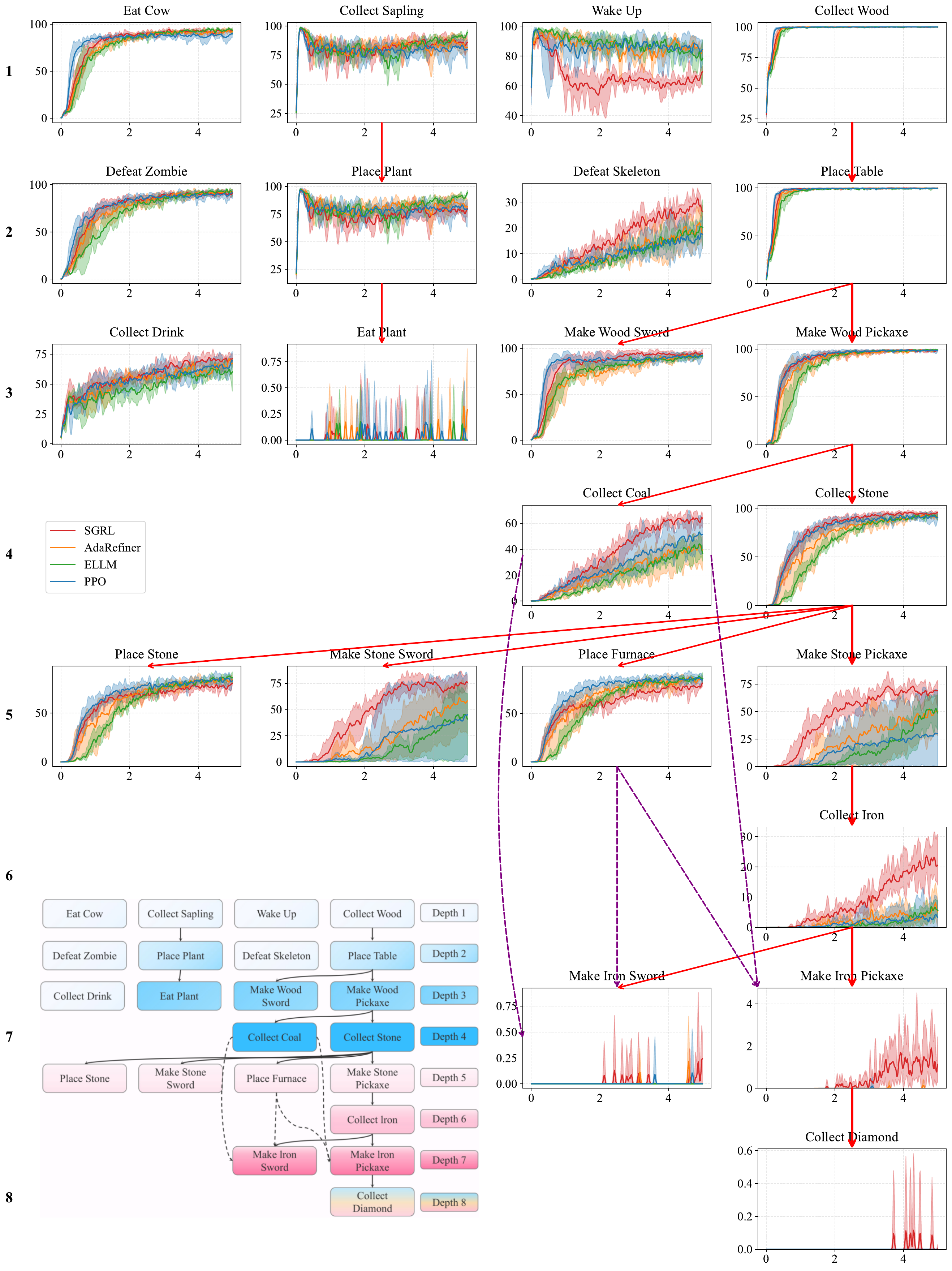}
	\caption{Main Results. 
	Success rate curves for all achievements on Craftax-Classic within 5M steps.
	Solid and dashed arrows indicate direct and cross-depth dependencies, respectively. 
	The bottom-left panel visualizes the full achievement dependency graph, with achievement depth encoded by color (depth 1-8 from top to bottom). 
	}
	\label{AppendixFig: fig_22ach_tax_baseline_5M}
\end{figure}

\begin{figure}[h]
	\vspace{2em}
	\centering
	\includegraphics[width=0.98\textwidth]{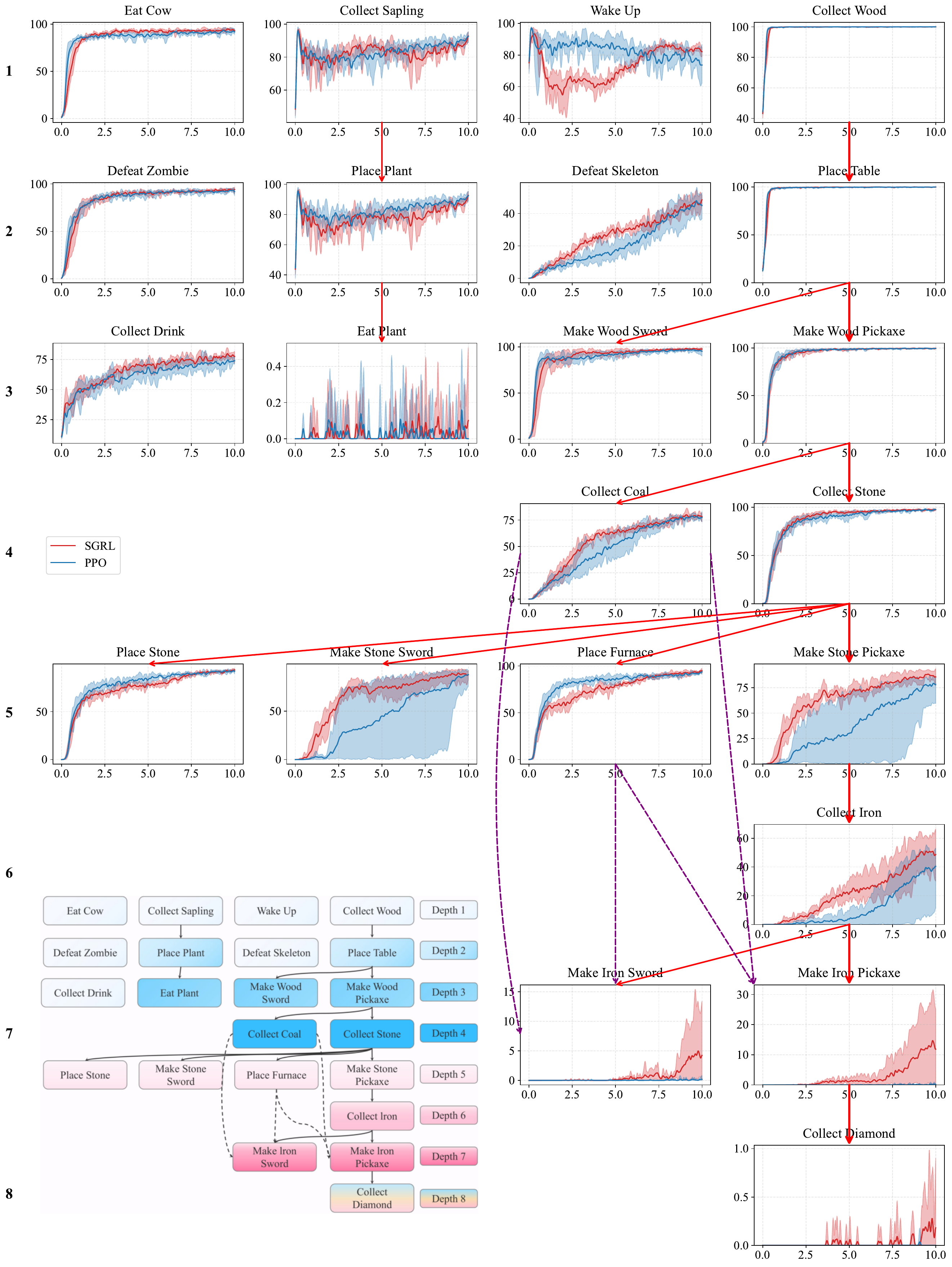}
	\caption{Main Results.	
	Success rate curves for all achievements on Craftax-Classic within 10M steps.
	Solid and dashed arrows indicate direct and cross-depth dependencies, respectively. 
	The bottom-left panel visualizes the full achievement dependency graph, with achievement depth encoded by color (depth 1-8 from top to bottom). 
	}
	\label{AppendixFig: fig_22ach_tax_baseline_10M}
	\vspace{2em}
\end{figure}

\begin{figure}[h]
	\centering
	\begin{subfigure}[b]{0.98\textwidth}
		\centering
		\includegraphics[width=\textwidth]{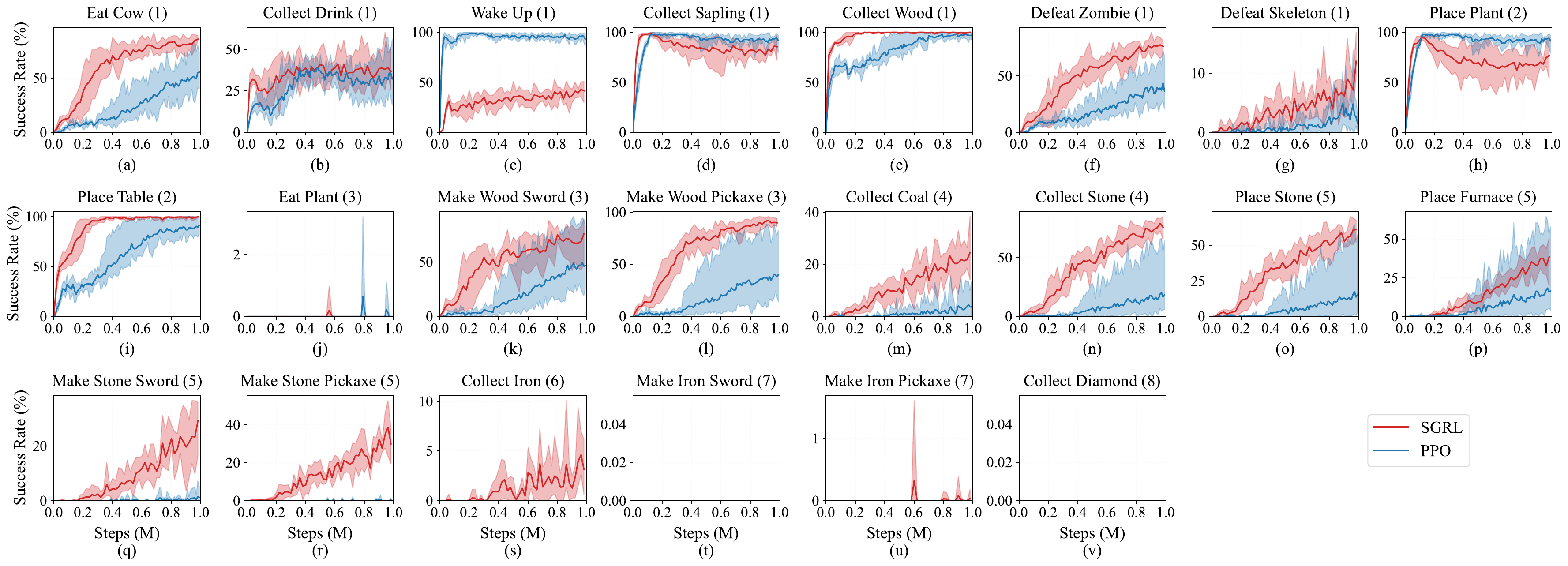} 
		\caption{1M Steps}
		\label{AppendixFig: fig_ter_beseline_1M}
	\end{subfigure}

	\begin{subfigure}[b]{0.98\textwidth}
		\centering
		\includegraphics[width=\textwidth]{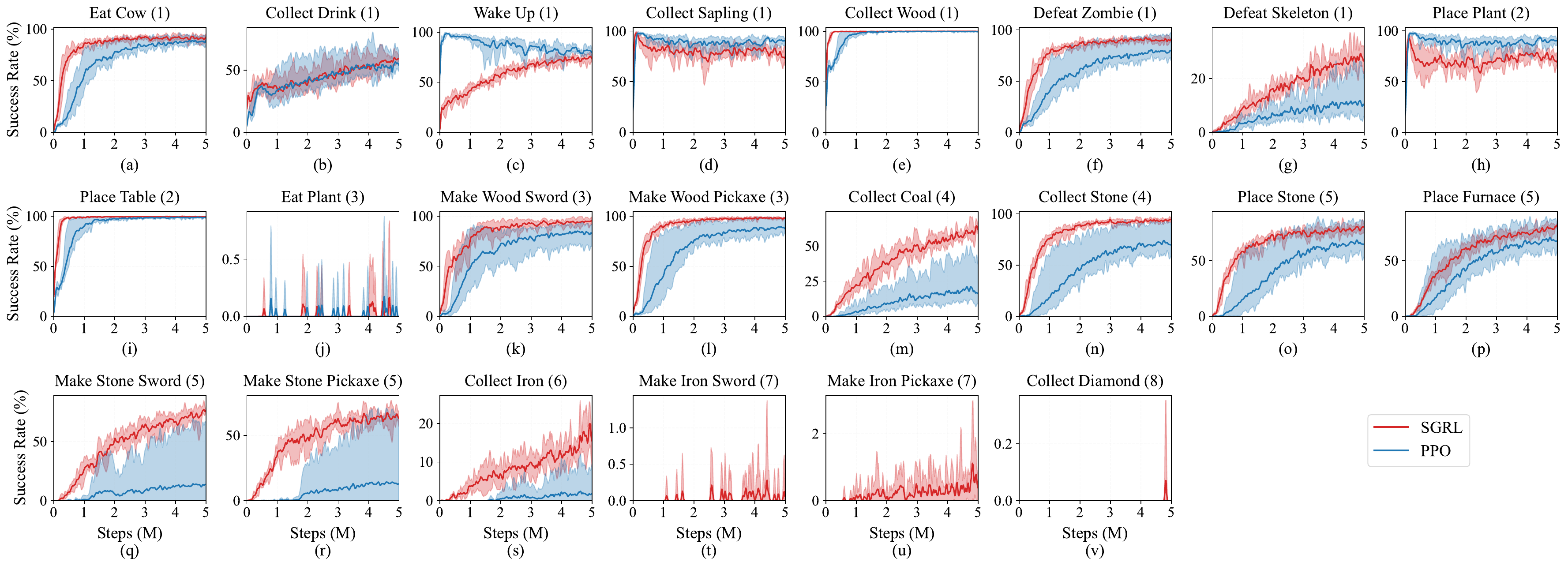} 
		\caption{5M Steps}
		\label{AppendixFig: fig_ter_beseline_5M}
	\end{subfigure}
	\caption{Main Results. 	
		Success rate curves for all achievements on Crafter within different training steps.
	We rank the achievements based on their depth and the importance of unlocking them for subsequent tasks. 
	Achievements ranked later have greater depth and exert a stronger influence on subsequent achievements.
	A more intuitive version is shown in Figure \ref{AppendixFig: fig_22ach_ter_baseline_5M}.
	}
	\label{AppendixFig: fig_ter_beseline}
\end{figure}

\begin{figure}[h]
	\centering
	\begin{subfigure}[b]{0.98\textwidth}
		\centering
		\includegraphics[width=\textwidth]{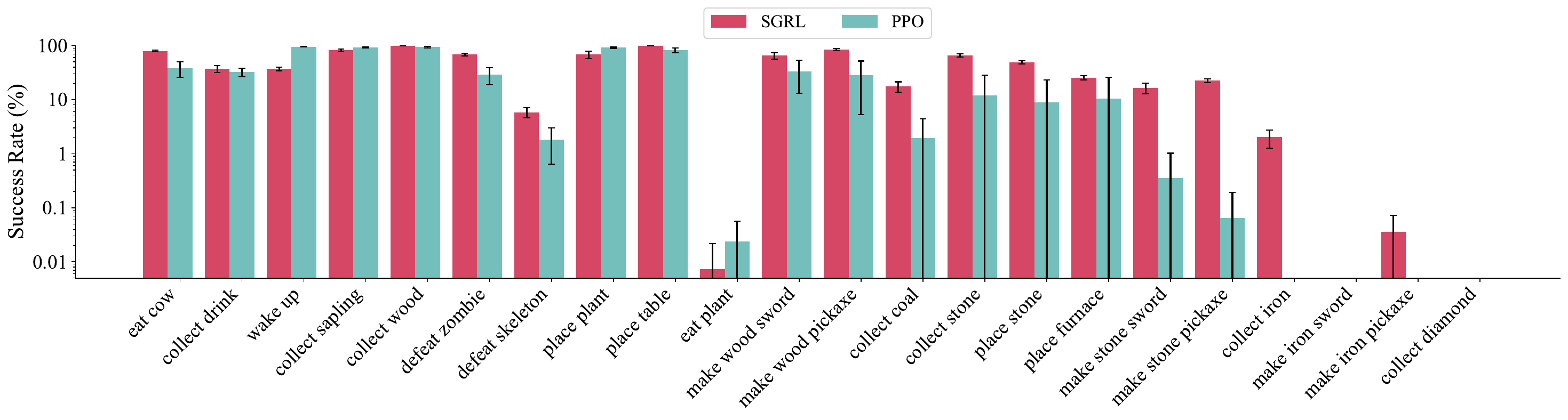} 
		\caption{1M Steps}
		\label{AppendixFig: fig_bar_ter_beseline_1M}
	\end{subfigure}
	\begin{subfigure}[b]{0.98\textwidth}
		\centering
		\includegraphics[width=\textwidth]{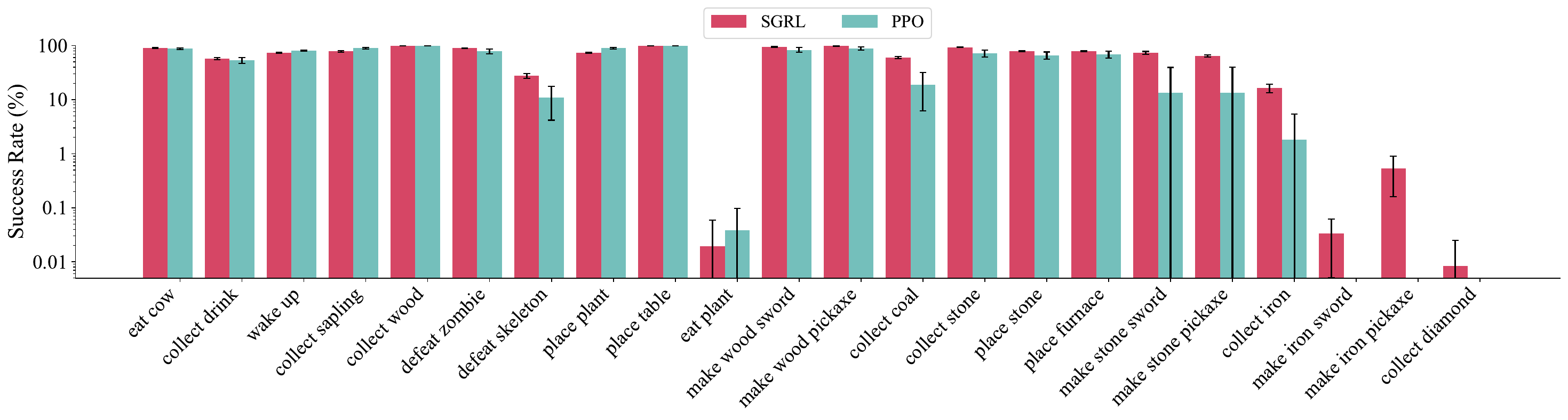} 
		\caption{5M Steps}
		\label{AppendixFig: fig_bar_ter_beseline_5M}
	\end{subfigure}
	\caption{Main Results. 
	Success rates across all achievements on Crafter at different training steps.
	}
	\label{AppendixFig: fig_bar_ter_beseline}
\end{figure}

\begin{figure}[h]
	\vspace{2em}
	\centering
	\includegraphics[width=0.98\textwidth]{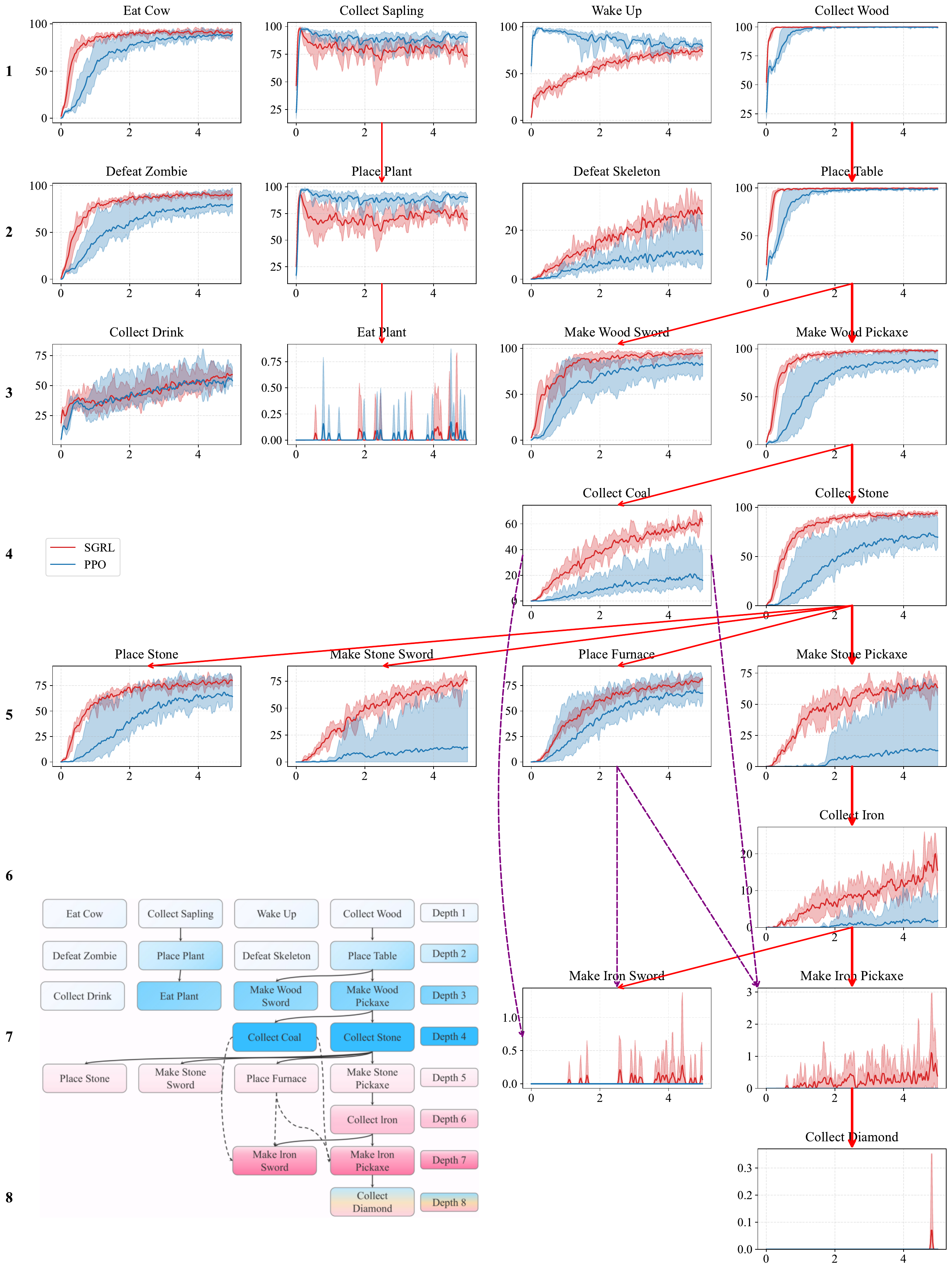}
	\caption{Main Results. 
		Success rate curves for all achievements on Crafter within 5M steps.
		Solid and dashed arrows indicate direct and cross-depth dependencies, respectively. 
		The bottom-left panel visualizes the full achievement dependency graph, with achievement depth encoded by color (depth 1-8 from top to bottom). 
	}
	\label{AppendixFig: fig_22ach_ter_baseline_5M}
	\vspace{2em}
\end{figure}

Table \ref{AppendixTab: crafter_perfor_5m} presents performance of SGRL and baseline methods on Crafter at 5M steps.
Figures \ref{AppendixFig: fig_tax_beseline}-\ref{AppendixFig: fig_22ach_ter_baseline_5M} present more detailed results of the main experiments on Crafter and Craftax-Classic.
It is worth noting that:
\begin{itemize}
	\item In the experiments on Craftax-Classic, ELLM and AdaRefiner require frequent online calls to the LLM (\texttt{DeepSeek-V3}) during training, incurring substantial computational costs and training time. 
	Therefore, we only reproduce the results within 5M steps.

	\item Since the Crafter environment does not adopt the JAX framework and runs extremely slowly, we report the original results of ELLM and AdaRefiner from their papers in Table~\ref{AppendixTab: crafter_perfor_5m}, rather than reproducing their experiments.
\end{itemize}

\textbf{Note:} Since ELLM and AdaRefiner require frequent online calls to the LLM (\texttt{DeepSeek-V3}) during training, they incur substantial computational costs and training time. 
Therefore, we only reproduce the results within 5M steps.

\section{Additional Ablation Experiments}  \label{Appendix E: Additional Ablation Experiments}
\subsection{Performance of Ablation Algorithm}   \label{Appendix E.1: Performance of Ablation Algorithm}
\begin{figure}[h]
	\vspace{2em}
	\centering
	\begin{subfigure}[b]{0.98\textwidth}
		\centering
		\includegraphics[width=\textwidth]{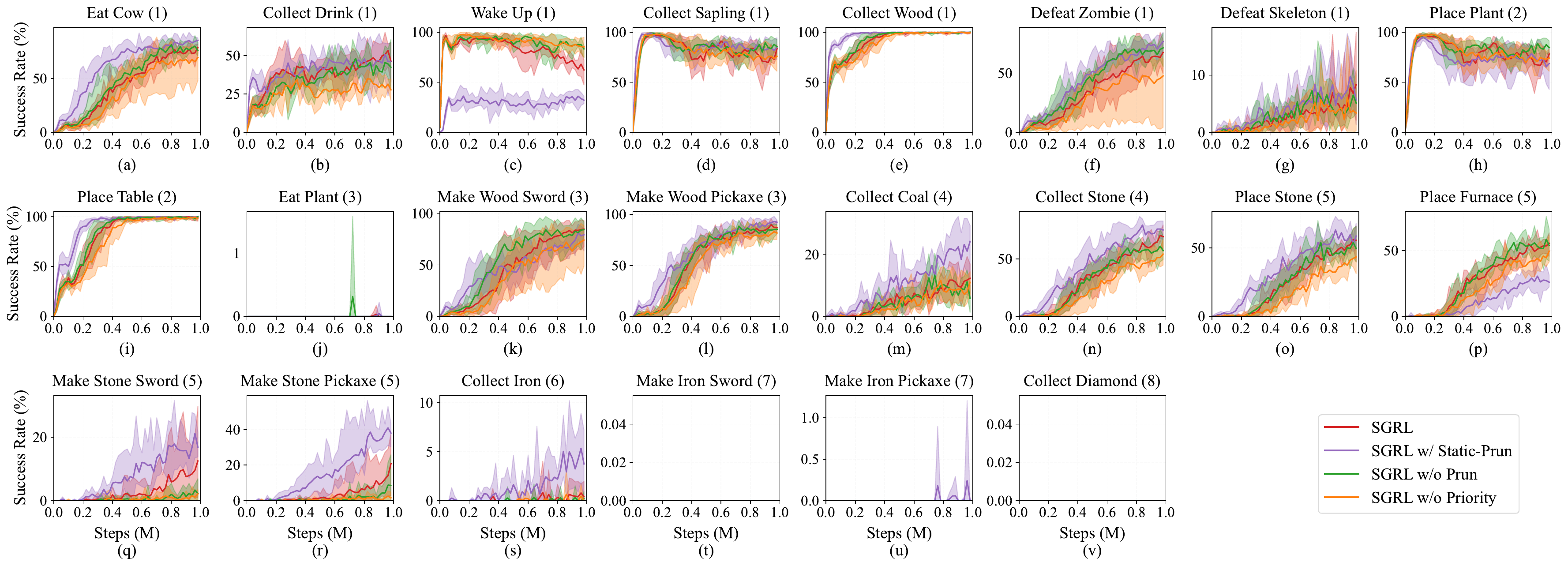} 
		\caption{1M Steps}
		\label{AppendixFig: fig_1M}
	\end{subfigure}

	\begin{subfigure}[b]{0.98\textwidth}
		\centering
		\includegraphics[width=\textwidth]{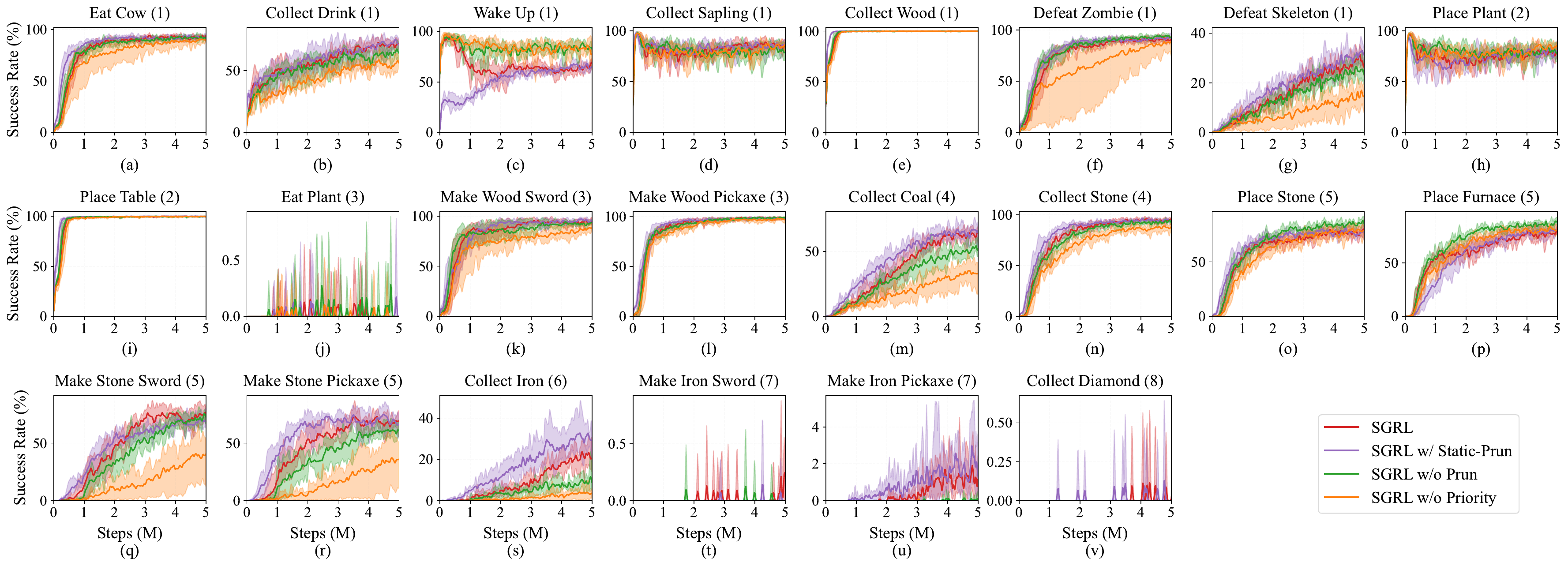} 
		\caption{5M Steps}
		\label{AppendixFig: fig_5M}
	\end{subfigure}

	\begin{subfigure}[b]{0.98\textwidth}
		\centering
		\includegraphics[width=\textwidth]{fig_10M.pdf} 
		\caption{10M Steps}
		\label{AppendixFig: fig_10M}
	\end{subfigure}
	\caption{Ablation Studies. 
	Success rate curves for all achievements on Craftax-Classic within different training steps.
	We rank the achievements based on their depth and the importance of unlocking them for subsequent tasks. 
	Achievements ranked later have greater depth and exert a stronger influence on subsequent achievements.
	A more intuitive version is shown in Figure \ref{AppendixFig: fig_22ach_tax_abl_10M}.
	}
	\label{AppendixFig: fig_tax}
	\vspace{2em}
\end{figure}

\begin{figure}[h]
	\centering
	\begin{subfigure}[b]{0.98\textwidth}
		\centering
		\includegraphics[width=\textwidth]{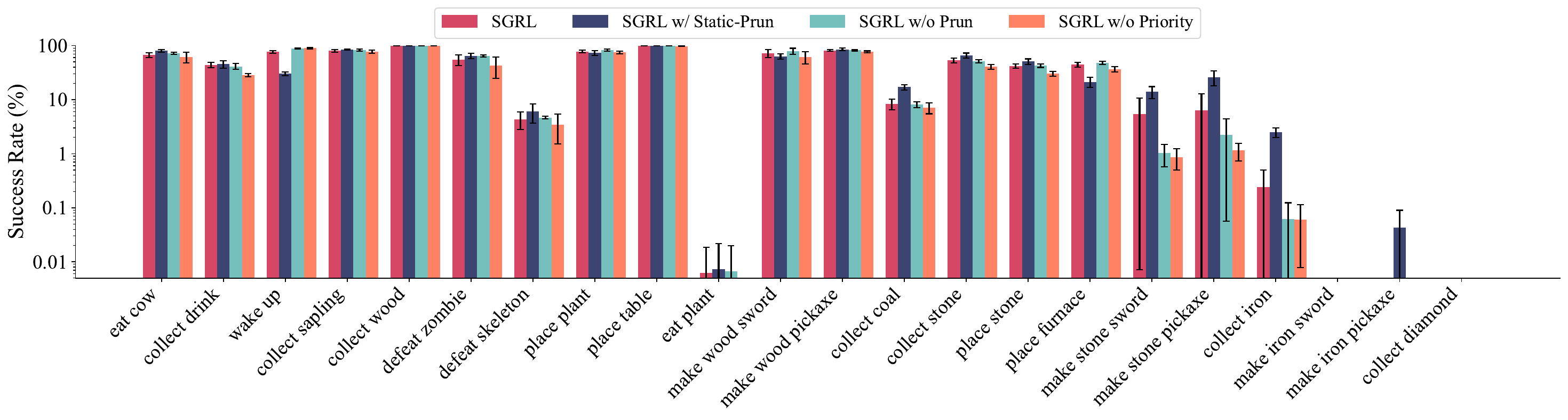} 
		\caption{1M Steps}
		\label{AppendixFig: fig_bar_1M}
	\end{subfigure}
	\begin{subfigure}[b]{0.98\textwidth}
		\centering
		\includegraphics[width=\textwidth]{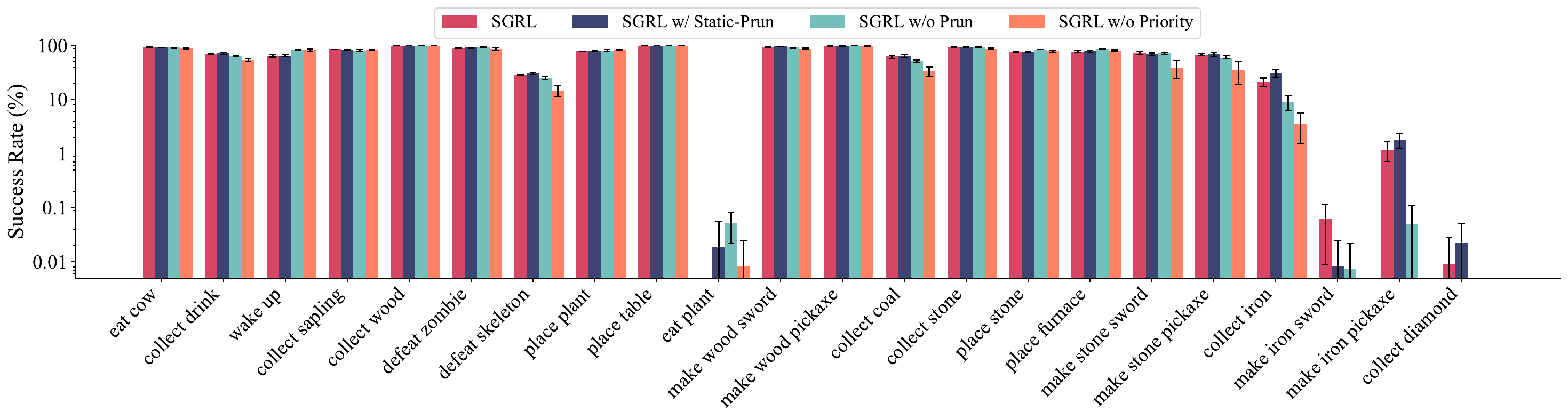} 
		\caption{5M Steps}
		\label{AppendixFig: fig_bar_5M}
	\end{subfigure}
	\begin{subfigure}[b]{0.98\textwidth}
		\centering
		\includegraphics[width=\textwidth]{fig_bar_10M.pdf} 
		\caption{10M Steps}
		\label{AppendixFig: fig_bar_10M}
	\end{subfigure}
	\caption{Ablation Studies. 
	Success rates across all achievements on Craftax-Classic at different training steps.
	}
    \label{AppendixFig: fig_bar}
\end{figure}

\begin{figure}[h]
	\centering
	\includegraphics[width=0.98\textwidth]{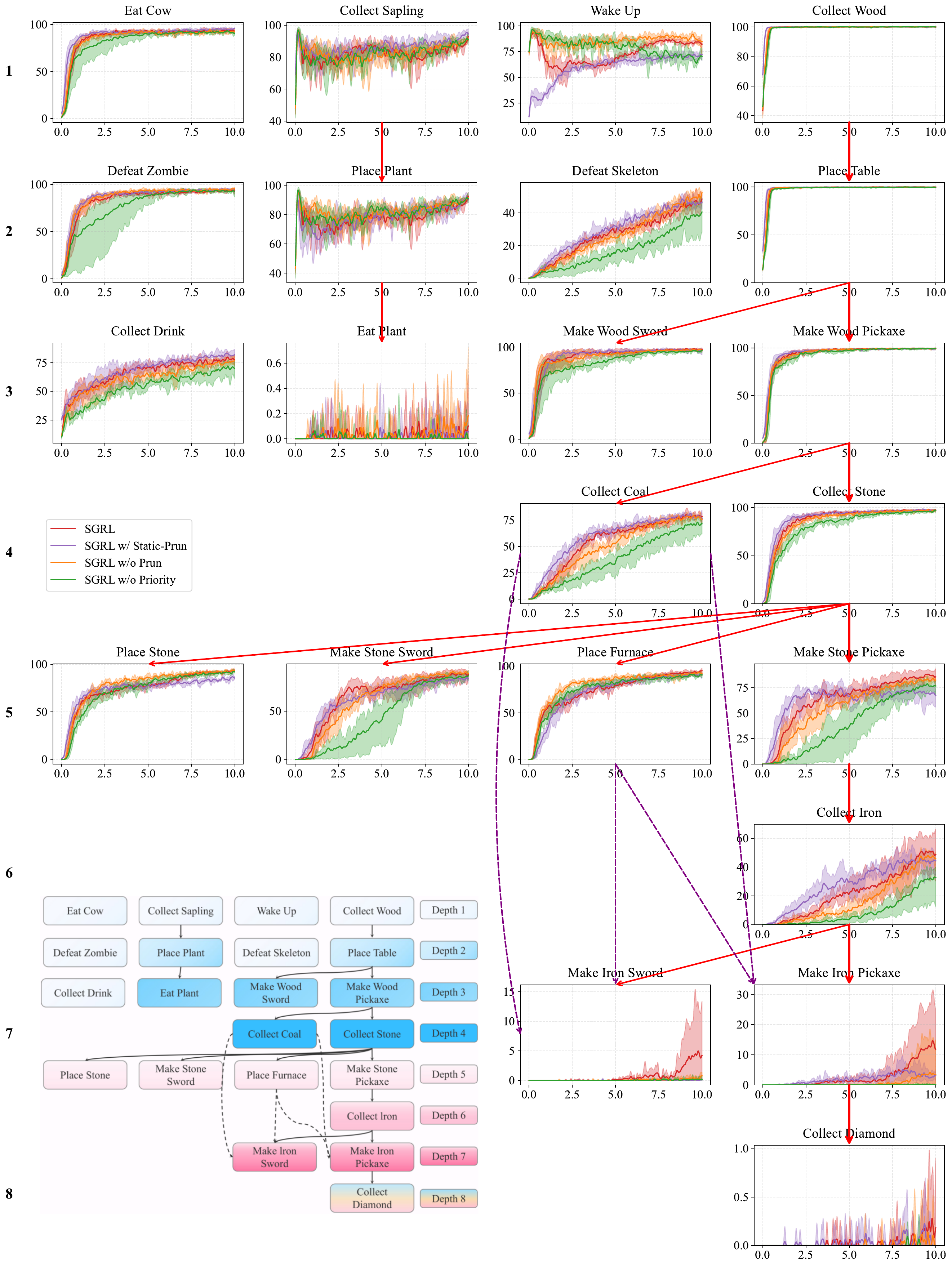}
	\caption{Ablation Studies. 
	    Success rate curves for all achievements on Craftax-Classic within 5M steps.
		Solid and dashed arrows indicate direct and cross-depth dependencies, respectively. 
		The bottom-left panel visualizes the full achievement dependency graph, with achievement depth encoded by color (depth 1-8 from top to bottom). 
	}
	\label{AppendixFig: fig_22ach_tax_abl_10M}
\end{figure}

Table~\ref{Tab: craftax_5m} and Figures~\ref{AppendixFig: fig_tax}-~\ref{AppendixFig: fig_22ach_tax_abl_10M} present more detailed results of the ablation experiments on Craftax-Classic.

\subsection{Heatmap of Goal with Priority Weights}   \label{Appendix E.2: Heatmap of Goal with Priority Weights}
\begin{figure}[h]
	\centering
	\includegraphics[width=0.98\textwidth]{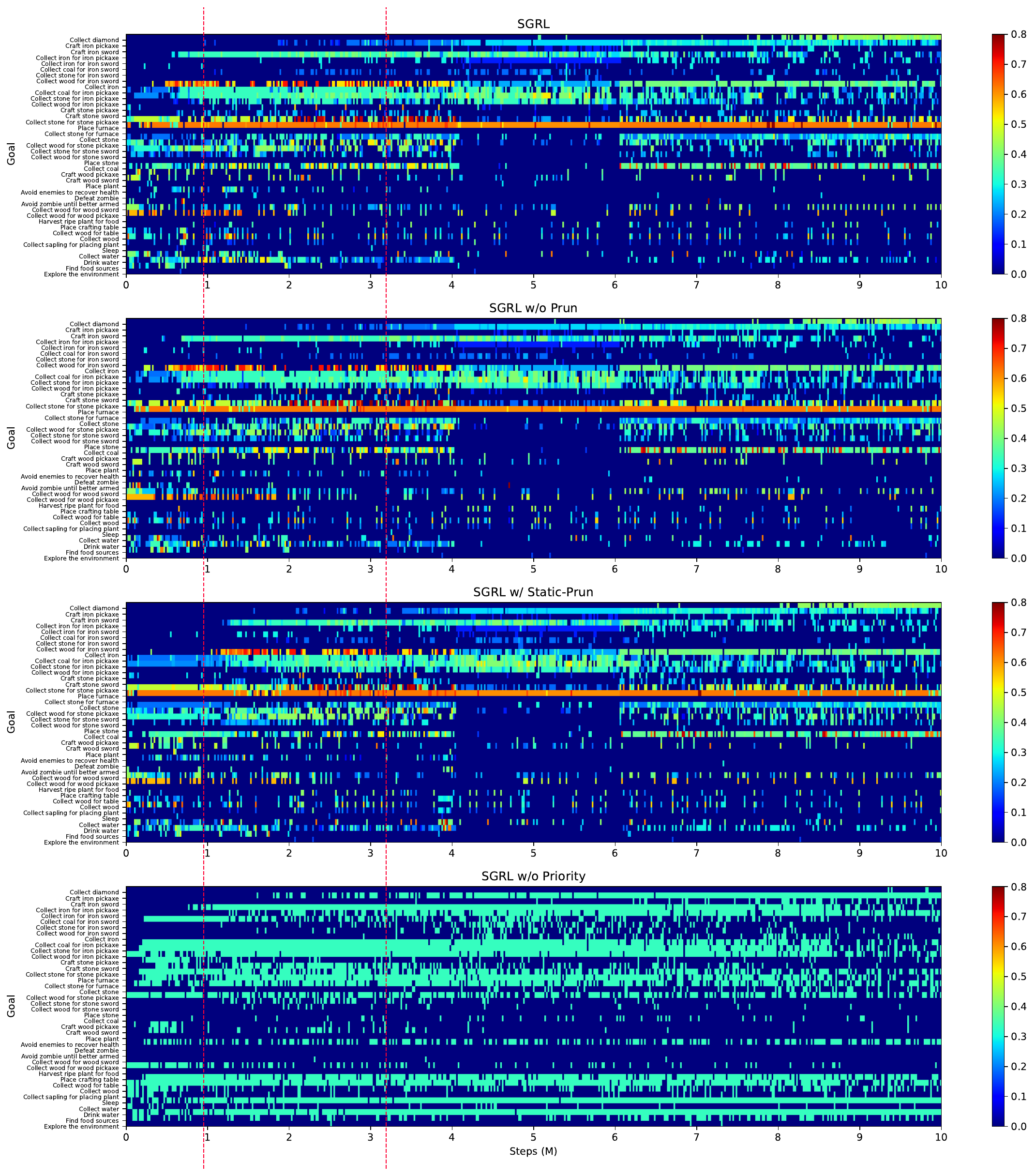}
	\caption{Ablation Studies. 
	Heatmap of the goals with priority weights generated by the structured goal planner on Craftax-Classic within 10M steps. 
	The vertical axis on the left shows goals ranked from low to high, while the right axis (ranging from 0 to 0.8) indicates the corresponding weights. 
	}
	\label{AppendixFig: heatmap_goal}
\end{figure}

Figure~\ref{AppendixFig: heatmap_goal} show the heatmap of the goals with priority weights generated by the structured goal planner on Craftax-Classic within 10M steps. 
The vertical axis on the left shows goals ranked from low to high, while the right axis (ranging from 0 to 0.8) indicates the corresponding weights.

From Figure~\ref{AppendixFig: heatmap_goal}, we can observe the following key phenomena:
\begin{itemize}
	\item SGRL w/o Priority only sets collect diamond as an exploration goal after 9M steps, which likely accounts for its significantly low exploration efficiency.
	\item SGRL, SGRL w/ Static Pruning, and SGRL w/o Pruning—methods that assign priority weights to goals—consistently treat long-horizon, high-impact achievements (e.g., Collect stone for stone pickaxe, Collect iron) as important objectives and assign them larger priority weights.
	\item In contrast to SGRL w/ Static Pruning and SGRL w/o Pruning, SGRL consistently assigns relatively small priority weights to more forward-looking goals (e.g., Collect diamond; see the red dashed lines in the figure), which encourages the agent to begin exploring these challenging achievements earlier.
        We hypothesize that this adaptive prioritization strategy is the primary driver behind SGRL's superior performance.
\end{itemize} 

\section{Additional Pruner Annealing Strategies and Implementation}  \label{Appendix F: Additional Pruner Epsilon Annealing Strategies and Implementation}
\subsection{Pruner Annealing Strategies}  \label{Appendix F.1: Pruner Annealing Strategies}
To dynamically adjust the agent's reliance on constraints during training, we implement several annealing strategies for $\xi$. 
These strategies smoothly modulate $\xi$ over training steps. 
These annealing schedules aim to balance the agent's adherence to the action pruner and the freedom of policy exploration.

\begin{itemize}
	\item \textit{Linear Annealing.}
 	This strategy linearly anneals $\xi$ from 0 to 1, gradually reducing the influence of the pruner.
	\begin{equation}
		\xi(t) = \frac{t}{T},
		\label{eq:linear}
	\end{equation}
	where $t$ is current training step, $T$ is total training steps.
	\item \textit{Exponential Annealing.}
	This strategy uses exponential decay to increase $\xi$ rapidly from 0 toward 1, then asymptotically approach full freedom.
	\begin{equation}
		\xi(t) = 1 - \exp\left(-\frac{t}{\tau}\right), 
		\label{eq:exp}
	\end{equation}
	where $\tau$ is the time constant of the exponential decay.
	\item \textit{Three-Stage Linear Annealing.}
	This strategy consists of three phases and is defined by the piecewise function:
	\begin{equation}
		\xi(t) =
		\begin{cases}
			1 - \dfrac{t}{0.4 T}, & t < 0.4T \\
			\dfrac{t}{0.4 T} - 1, & 0.4T \leq t < 0.8T \\
			1, & t \geq 0.8T
		\end{cases}.
		\label{eq:3stage_linear}
	\end{equation}
	\item \textit{Three-Stage Cosine Annealing.} This strategy uses a smooth cosine function for the first two stages:
	\begin{equation}\renewcommand{\arraystretch}{1.5}
    \xi(t)=\left\{\begin{array}{ll}
        \frac{1}{2}\left(1+\cos \left(\frac{t}{0.4 T} \cdot \pi \right)\right),       & 0 \leq t<0.4 T     \\
        \frac{1}{2}\left(1-\cos \left(\frac{t-0.4 T}{0.4 T} \cdot \pi \right)\right), & 0.4 T \leq t<0.8 T \\
        1.0,                                                                          & t \geq 0.8 T
    \end{array}\right.
    .
    \label{eq:3stage_cos}
\end{equation}
\end{itemize}

In our experiments, the following naming convention is used to denote different annealing strategies applied to the SGRL:
(1) \textit{SGRL w/ Linear Ann}: SGRL equipped with linear annealing schedule (see Equation (\ref{eq:linear}));
(2) \textit{SGRL w/ Exp Ann}: SGRL with exponential annealing (see Equation (\ref{eq:exp}));
(3) \textit{SGRL w/ 3-Stage Linear}: SGRL with piecewise linear three-phase annealing (see Equation (\ref{eq:3stage_linear}));
and (4) \textit{SGRL w/ 3-Stage Cos}: SGRL with cosine-based three-phase annealing (see Equation (\ref{eq:3stage_cos})).

\subsection{Experimental Results} \label{Appendix F.2: Experimental Results}
\begin{figure}[h]
	\centering
	\begin{subfigure}[b]{0.98\textwidth}
		\centering
		\includegraphics[width=\textwidth]{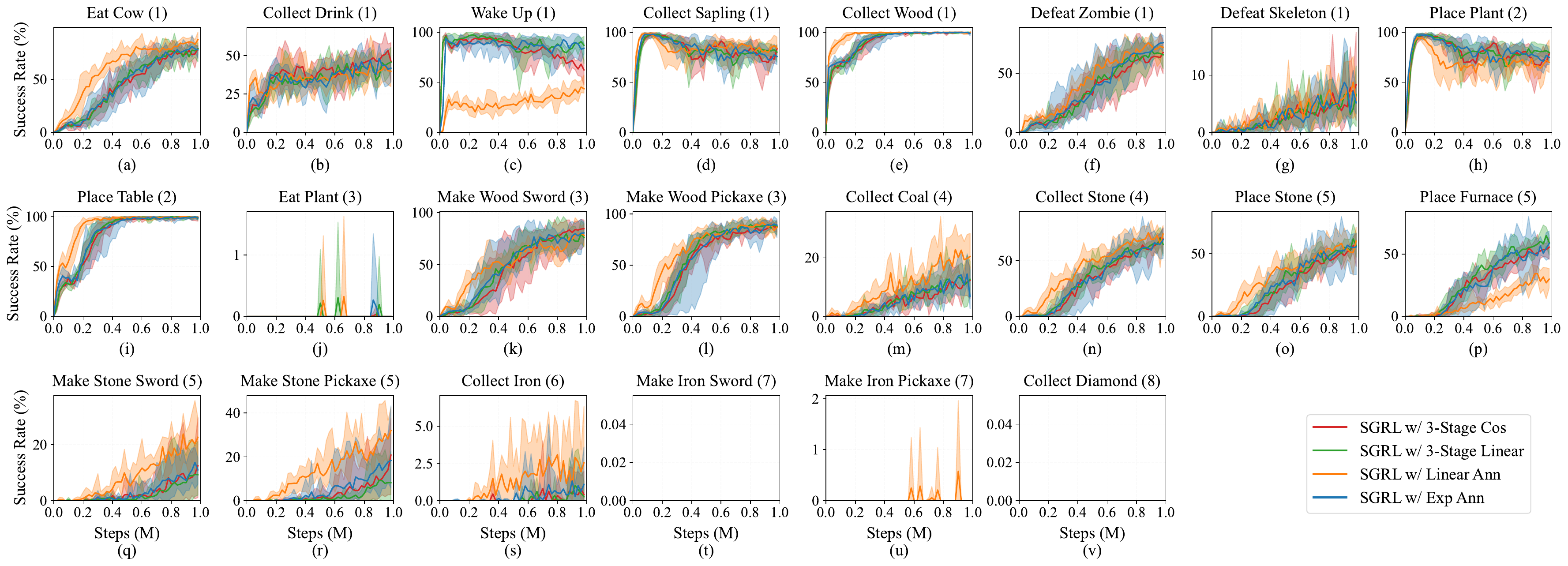} 
		\caption{1M Steps}
		\label{AppendixFig: fig_mask_1M}
	\end{subfigure}

	\begin{subfigure}[b]{0.98\textwidth}
		\centering
		\includegraphics[width=\textwidth]{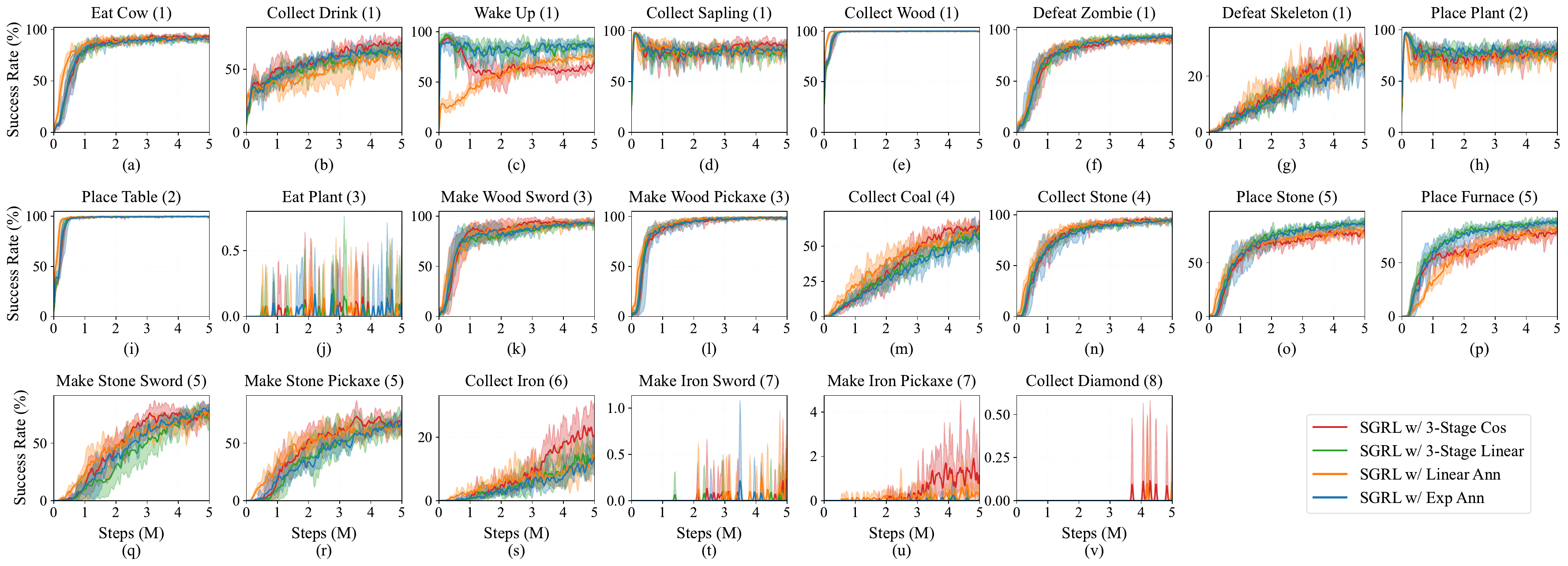} 
		\caption{5M Steps}
		\label{AppendixFig: fig_mask_5M}
	\end{subfigure}

	\begin{subfigure}[b]{0.98\textwidth}
		\centering
		\includegraphics[width=\textwidth]{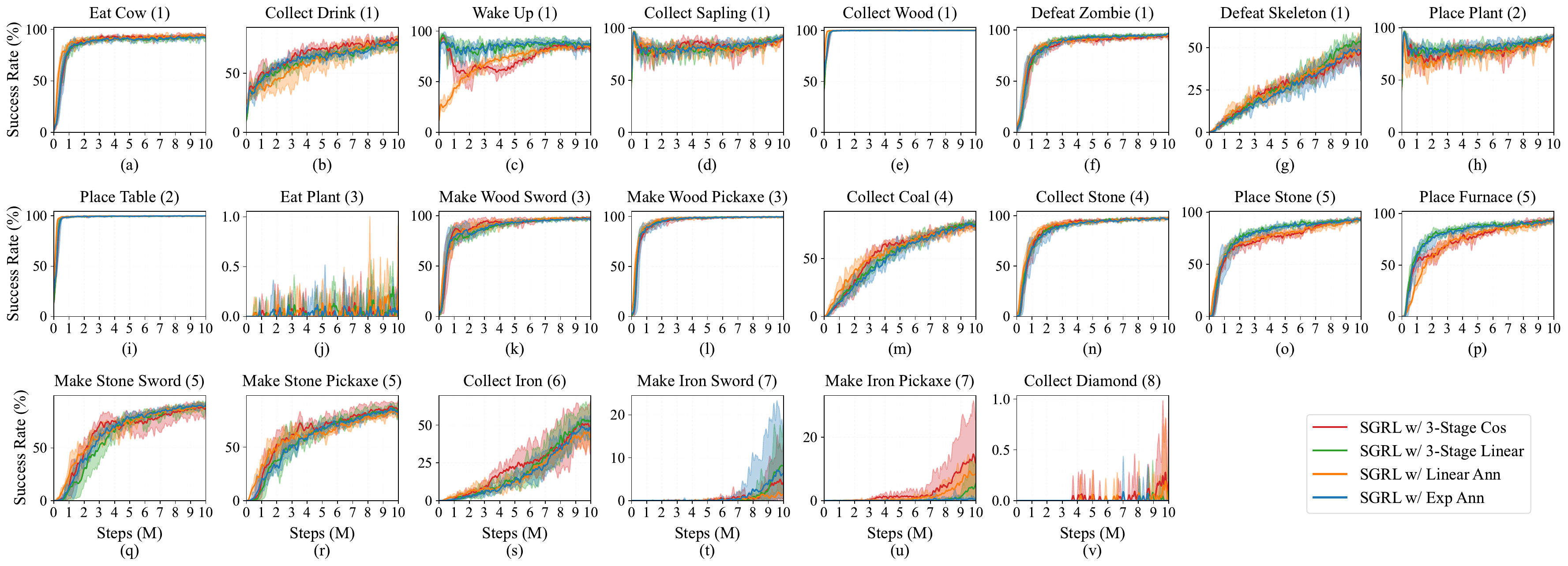} 
		\caption{10M Steps}
		\label{AppendixFig: fig_mask_10M}
	\end{subfigure}
	\caption{Mask Comparison.
	Success rate curves for all achievements on Craftax-Classic within different training steps.
	We rank the achievements based on their depth and the importance of unlocking them for subsequent tasks. 
	Achievements ranked later have greater depth and exert a stronger influence on subsequent achievements.
	A more intuitive version is shown in Figure \ref{AppendixFig: fig_22ach_tax_mask_10M}.
    }
	\label{AppendixFig: fig_tax_mask}
\end{figure}

\begin{figure}[h]
	\centering
	\begin{subfigure}[b]{0.98\textwidth}
		\centering
		\includegraphics[width=\textwidth]{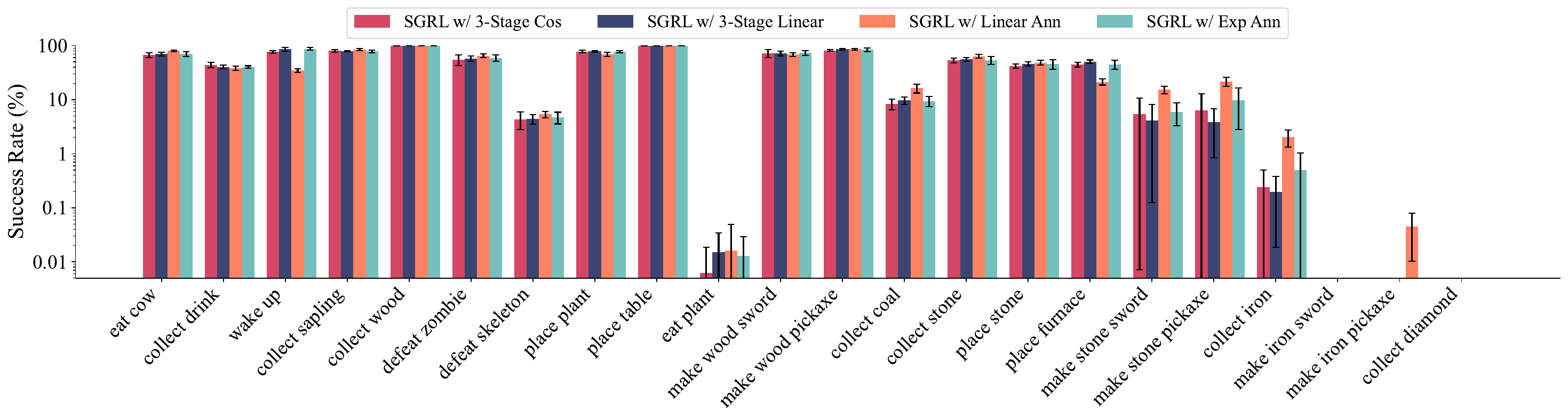} 
		\caption{1M Steps}
		\label{AppendixFig: fig_bar_mask_1M}
	\end{subfigure}
	\begin{subfigure}[b]{0.98\textwidth}
		\centering
		\includegraphics[width=\textwidth]{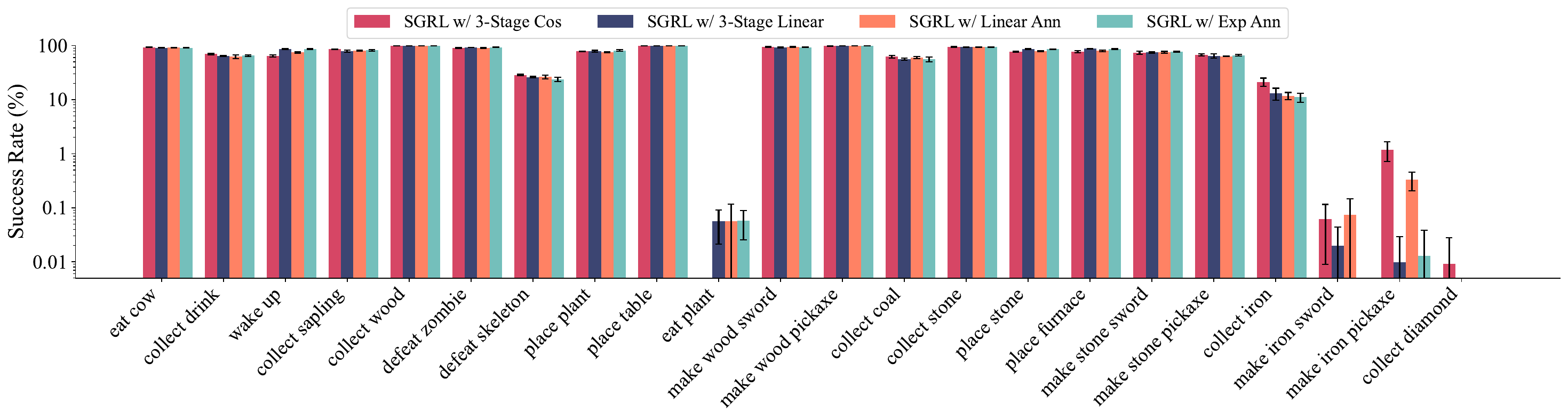} 
		\caption{5M Steps}
		\label{AppendixFig: fig_bar_mask_5M}
	\end{subfigure}
	\begin{subfigure}[b]{0.98\textwidth}
		\centering
		\includegraphics[width=\textwidth]{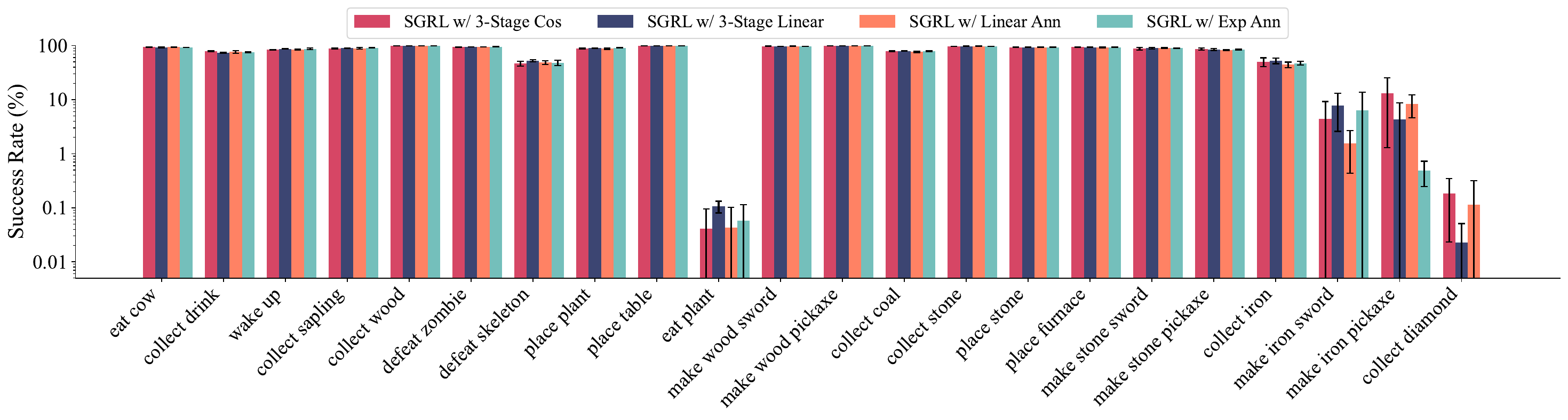} 
		\caption{10M Steps}
		\label{AppendixFig: fig_bar_mask_10M}
	\end{subfigure}
	\caption{Mask Comparison.
	Success rates across all achievements on Craftax-Classic at different training steps.
	 }
    \label{AppendixFig: fig_bar_mask}
\end{figure}

\begin{figure}[h]
	\centering
	\includegraphics[width=0.98\textwidth]{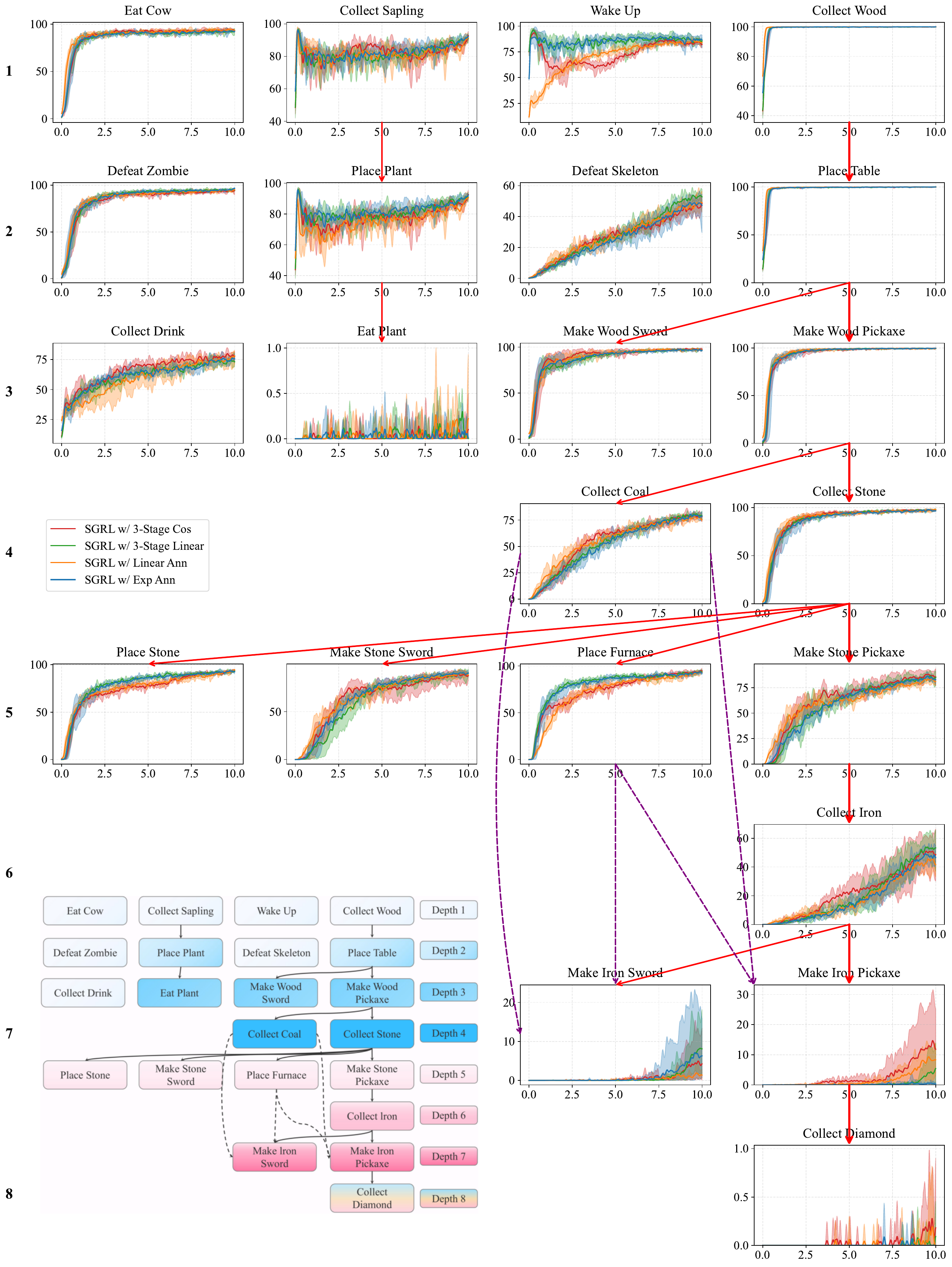}
	\caption{Mask Comparison.
		Success rate curves for all achievements on Craftax-Classic within 5M steps.
		Solid and dashed arrows indicate direct and cross-depth dependencies, respectively. 
		The bottom-left panel visualizes the full achievement dependency graph, with achievement depth encoded by color (depth 1-8 from top to bottom). 
	}
	\label{AppendixFig: fig_22ach_tax_mask_10M}
\end{figure}

Figures~\ref{AppendixFig: fig_tax_mask}-\ref{AppendixFig: fig_22ach_tax_mask_10M} present detailed results with different mask mechanism on Craftax-Classic.

As shown in Figures~\ref{AppendixFig: fig_tax_mask}-\ref{AppendixFig: fig_22ach_tax_mask_10M}, the performance of SGRL with four different $\xi$ annealing strategies on Craftax-Classic is presented:  
\begin{itemize}
	\item Figure~\ref{AppendixFig: fig_tax_mask} displays the success rate curves across 22 achievements on Craftax-Classic, comparing four mask annealing strategies. 
	Notably, SGRL w/ 3-Stage Cos achieves diamond collection at 3.7M steps and maintains superior performance on the most challenging achievements (Make Iron Pickaxe and Collect Diamond), as shown in Figures~\ref{AppendixFig: fig_tax_mask} (b)-(c). 
	This suggests that the Three-Stage Cosine Annealing strategy enables SGRL to more effectively prioritize high-value, long-horizon objectives by adaptively balancing exploration and exploitation.
	In contrast, SGRL w/ Linear Ann demonstrates stronger early-stage performance, unlocking depth-7 achievements faster (see Figure~\ref{AppendixFig: fig_tax_mask} (a)).
	However, its success rate plateaus in later training phases (see Figures~\ref{AppendixFig: fig_tax_mask} (b)-(c)), likely due to the rigid linear decay of $\xi$, which prematurely restricts exploration and hinders adaptation to complex tasks.  
	A success rate plot that intuitively reflects achievement depth is shown in Figure~\ref{AppendixFig: fig_22ach_tax_mask_10M}.

	\item Figure~\ref{AppendixFig: fig_bar_mask} presents the success rates across all 22 achievements on Craftax-Classic at different training steps. 
	From Figure~\ref{AppendixFig: fig_bar_mask}, we can observe that while \textit{SGRL w/ 3-Stage Linear} and \textit{SGRL w/ 3-Stage Cos} exhibit similar performance early on (Figure~\ref{AppendixFig: fig_bar_mask} (a)), the latter significantly outperforms the former in late-stage deep achievements (see Figures~\ref{AppendixFig: fig_bar_mask} (b)-(c)). 
	We hypothesize that the piecewise linear transitions in 3-Stage Linear Annealing introduce abrupt changes in exploration pressure, whereas the smooth cosine modulation in 3-Stage Cosine Annealing facilitates more stable learning.
	Interestingly, \textit{SGRL w/ Exp Ann} is the only variant failing to unlock Collect Diamond by 10M steps. 
	This indicates that the rapid decay of $\xi$ in exponential annealing diminishes goal guidance too early, impairing the agent's ability to align goals with agents' actions and hindering the acquisition of complex, multi-step behaviors.
	
\end{itemize}
Overall, these results highlight the critical role of the annealing schedule in modulating the trade-off between goal-driven exploration and policy autonomy. 
The three-stage cosine strategy achieves the most effective balance, enabling sustained guidance during critical phases of skill acquisition while allowing gradual transition to policy-based control.

\begin{table}[h]
	\centering
	\setlength{\tabcolsep}{4pt}
	\begin{threeparttable}
		\begin{tabular}{l c c c c}
			 \toprule
             \textbf{Method} & \textbf{Score(\%)} & \textbf{Reward} & \makecell{\textbf{Achievement Depth}} &  \textbf{SPS ($\times 10^2$)}  \\
			 \midrule
			 Human      & 50.5 $\pm$ 6.8   & 14.3 $\pm$ 2.3   & 8 & -  \\
			 \hline
			 SGRL       & \textbf{30.5} $\pm$ 1.2   & 12.7 $\pm$ 0.4   & \textbf{8} & 1.0  \\
			 AdaRefiner & 28.2 $\pm$ 1.8   & \textbf{12.9} $\pm$ 1.2   & 7 & -  \\
			 ELLM       & -                & 6.0  $\pm$ 0.4   & - & - \\
			 PPO        & 18.5 $\pm$ 6.1   & 10.1 $\pm$ 1.3   & 6 & 1.2 \\
			 \bottomrule
		\end{tabular}	
	\end{threeparttable}
	\caption{Main Results.
		Performance of SGRL and baseline methods on Crafter at 5M steps.}
	\label{AppendixTab: crafter_perfor_5m}
	\vspace{2em}
 \end{table}

\begin{table}[h]
	\centering
	\setlength{\tabcolsep}{4pt}
	\begin{threeparttable}
		\begin{tabular}{l c c c c}
			 \toprule
             \textbf{Method} & \textbf{Score(\%)} & \textbf{Reward} & \makecell{\textbf{Achievement Depth}}  \\
			 \midrule
			 SGRL                  & $33.8 \pm 1.5$           & $\textbf{13.0}\pm 0.3$ & 8                              \\
			 SGRL w/ Static-Prun   & $\textbf{34.2} \pm 1.7 $ & $12.9 \pm 0.4$         & 8                              \\
			 SGRL w/o Prun         & $30.9 \pm 1.3$           & $12.7 \pm 0.2$         & 7                              \\
			 SGRL w/o Priority     & $30.4 \pm 0.9$          & $12.3 \pm 0.5$          & 7                              \\
			 \bottomrule
		\end{tabular}	
	\end{threeparttable}
	\caption{Ablation Studies. 
	Performance of SGRL and ablation methods on Craftax-Classic at 5M steps.}
	\label{Tab: craftax_5m}
	\vspace{2em}
 \end{table}

\end{document}

%% file: math_commands.tex
\usepackage{amsmath,amsfonts,bm}

\def\eqref#1{equation~\ref{#1}}

\def\1{\bm{1}}

\DeclareMathAlphabet{\mathsfit}{\encodingdefault}{\sfdefault}{m}{sl}
\SetMathAlphabet{\mathsfit}{bold}{\encodingdefault}{\sfdefault}{bx}{n}

